\documentclass{article}

\usepackage[T1]{fontenc}
\usepackage[utf8]{inputenc}
\usepackage[letterpaper,margin=1in]{geometry}
\usepackage{times}
\usepackage[authoryear,round]{natbib}
\usepackage{fancyhdr}
\fancypagestyle{plain}{\fancyhf{}\fancyhead[L]{\small Under review}\fancyfoot[C]{\thepage}}
\usepackage{amsmath,amsfonts,bm}

\def\eqref#1{equation~\ref{#1}}

\def\1{\bm{1}}

\DeclareMathAlphabet{\mathsfit}{\encodingdefault}{\sfdefault}{m}{sl}
\SetMathAlphabet{\mathsfit}{bold}{\encodingdefault}{\sfdefault}{bx}{n}

\newcommand{\E}{\mathbb{E}}

\newcommand{\KL}{D_{\mathrm{KL}}}

\usepackage{amssymb}
\usepackage{amsthm}
\usepackage{mathtools}

\usepackage{wrapfig}
\usepackage{graphicx}
\usepackage{booktabs}
\usepackage{tabularx}
\usepackage{array}
\usepackage{multirow}
\usepackage{longtable}
\usepackage{enumitem}
\usepackage{float}
\usepackage{algorithm}
\usepackage{algpseudocode}

\usepackage[table]{xcolor}
\usepackage{microtype}

\usepackage{hyperref}
\usepackage{url}
\usepackage[capitalize,noabbrev]{cleveref}

\definecolor{TheoryBlue}{HTML}{174A7E}
\definecolor{PaleBlue}{HTML}{EEF5FB}

\hypersetup{
    colorlinks=true,
    linkcolor=TheoryBlue,
    citecolor=TheoryBlue,
    urlcolor=TheoryBlue
}

\newcommand{\mainblue}[1]{\textcolor{TheoryBlue}{#1}}

\newcommand{\sg}{\operatorname{sg}}

\newcommand{\norm}[1]{\left\lVert #1 \right\rVert}
\newcommand{\ip}[2]{\left\langle #1,#2 \right\rangle}

\newcommand{\eid}[1]{%
    \hyperref[sec:evidence-register]{\textsf{[#1]}}%
}

\newcolumntype{Y}{>{\raggedright\arraybackslash}X}

\theoremstyle{plain}
\newtheorem{proposition}{Proposition}

\theoremstyle{definition}
\newtheorem{assumption}{Assumption}

\theoremstyle{remark}

\setlist[itemize]{
    leftmargin=*,
    topsep=2pt,
    itemsep=1pt
}

\title{
Functional Cross-layer Experts Routing for \\
One-step Diffusion Distillation
}

\author{
  Zile Huang \quad Ser-nam Lim \\
  University of Central Florida \\
  \texttt{zi057775@ucf.edu} \quad \texttt{sernam@ucf.edu}
}

\date{}

\begin{document}

\maketitle

\begin{abstract}
One-step distillation compresses diffusion trajectories into a single forward evaluation, yet how internal capacity should be structured within that remaining pass remains largely unexplored. In this setting, identical parameters must simultaneously preserve general generative priors and resolve prompt-specific details without iterative refinement. We formalize this challenge as \emph{functional allocation}. Our analysis reveals that decoupling representations is beneficial only when expert interference outweighs shared transfer, favoring an always-active shared backbone augmented with conditionally routed residuals over standard isolated MoE architectures. We propose \textbf{DeMoE}, which pairs a persistent shared expert with conditionally routed expert slices across image FFNs, trained under a one-step distribution-matching objective with executed-route supervision. On one-step FLUX.2-4B generation, DeMoE achieves a GenEval score of 0.8220 while activating only 80.5\% of the image-FFN width---substantially outperforming dense one-step baselines, matching the 50-step teacher overall, and surpassing it on complex composition and counting. Across COCO-10K, DeMoE establishes state-of-the-art fidelity and diversity among single-step models, demonstrating that structured internal computation reallocation provides a superior compression paradigm for diffusion models. Code is available at \url{https://github.com/toffeecat/demoe-ttt-router}.
\end{abstract}

\section{Introduction}
\label{sec:introduction}

Diffusion models~\citep{ho2020denoising,song2020score,dhariwal2021diffusionmodelsbeatgans} have emerged as a dominant paradigm for generative modeling, advancing from discrete-time formulations to continuous-time score SDEs and deterministic ODE sampling~\citep{song2020ddim}. While high-resolution synthesis is substantially facilitated by operating on compressed latent spaces~\citep{rombach2022high} and rectifying trajectories via flow matching~\citep{lipman2022flow,liu2022flow}, iterative generation remains fundamentally constrained by inference latency. Consequently, one-step generation tasks a single forward pass with what iterative samplers previously accomplished over dozens of denoiser evaluations. 

Despite this radical compression, that single pass remains a deep network: it must parse the prompt, compose objects, bind attributes, and render fine details before outputting an image, all without the opportunity to revisit or refine its predictions across consecutive diffusion steps. Compressing the sampling trajectory thus exposes a fundamental yet overlooked question: how should the computation preserved within this single pass be organized? \Cref{fig:method} (a) illustrates the structural differences among the three paradigms: the multi-step teacher evaluates its entire architecture iteratively, a dense one-step student evaluates it once, and DeMoE preserves the single forward pass while making internal computational pathways conditionally selectable.

Distribution Matching Distillation (DMD) serves as our starting point because it explicitly defines the learning objective for one-step students: it perturbs the generated endpoint and matches the student marginal to the teacher marginal at that perturbation, rather than forcing the student to reconstruct intermediate trajectory states~\citep{yin2024onestep,yin2024improved}. However, optimizing a distribution-matching objective only constrains the terminal mapping, leaving the student's internal capacity and execution pathways fundamentally flexible.
In parallel, Transformers have increasingly replaced traditional convolutional U-Nets as scalable generative backbones~\citep{peebles2023scalable,ma2024sit,Wang2025DDT}, wherein representation alignment strategies further accelerate training convergence~\citep{yu2024repa}. While sparse mixture-of-experts (MoE) and shared-expert paradigms already enable conditional execution~\citep{shazeer2017outrageously,fedus2022switch,dai2024deepseekmoe}, and diffusion transformers have dynamically routed tokens, denoising tasks, or generation phases~\citep{fei2024scaling,sun2024ec,park2024switchdit,fang2024remixdit,park2023dtr,Wang2025DDT}, our focus is more targeted: once a one-step student's forward pass is partitioned into discrete pathways, which capabilities must remain shared, which variations warrant conditional branching, and what runtime signals should steer routing?

Simply partitioning a feed-forward network (FFN) does not guarantee effective conditional computation; the bank must maintain specialized responses that benefit the current input, \emph{and} the controller must reliably dispatch them. We therefore decouple intrinsic capability from runtime allocation. Our controlled studies confirm that when all experts are simultaneously active, the bank does not collapse into redundant replicates; furthermore, under a fixed computational budget, selecting alternative pathways yields distinct generative behaviors.

\begin{figure}[t]
  \centering
  \includegraphics[width=\linewidth]{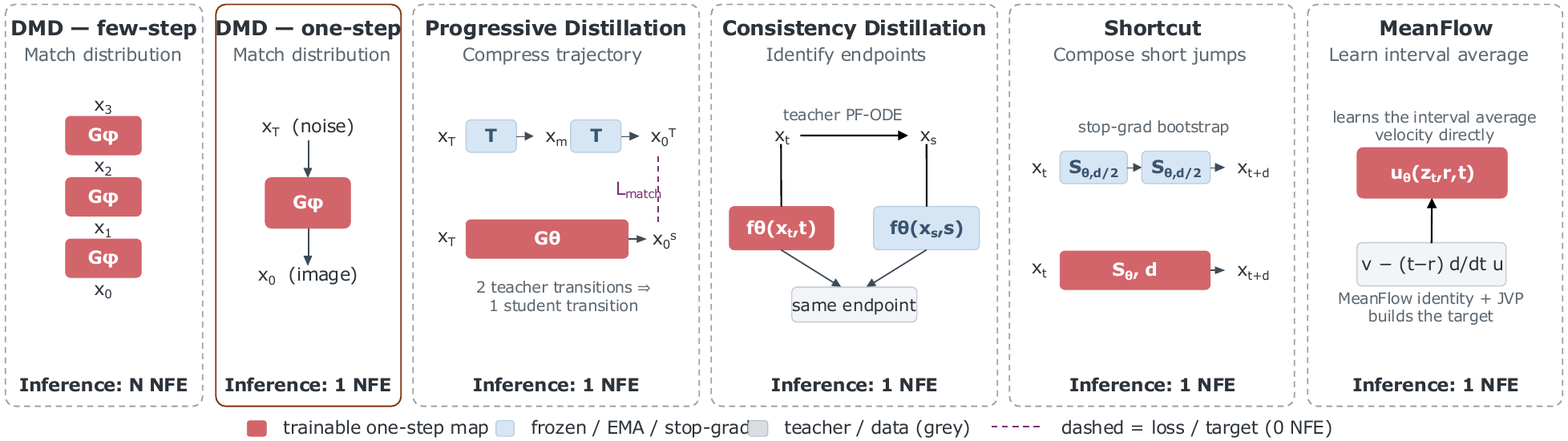}
\caption{
\textbf{Alternative ways to compress generative computation to one neural function evaluation.}
Few-step DMD retains an explicit sequence of state-dependent generator evaluations, whereas the remaining paradigms target one-step inference but introduce different forms of training structure.
One-step DMD removes the explicit sampling trajectory and matches the target distribution directly.
Progressive Distillation compresses multiple teacher transitions into one student transition;
Consistency Distillation transfers cross-time endpoint consistency;
Shortcut Models construct long transitions by composing shorter stop-gradient shortcuts;
and MeanFlow directly parameterizes the average velocity over a finite interval.
Thus, these methods differ primarily in \emph{how the one-step mapping is supervised}, even when their deployed inference graph contains only one network evaluation.
Red denotes the trainable one-step mapping, blue denotes frozen/EMA/stop-gradient computation, gray denotes teacher or data quantities, and dashed arrows denote losses or target construction.
}
  \label{fig: paradigm}
\end{figure}

A local interaction analysis provides the foundational design rule for our proposed architecture. When distinct input groups induce conflicting first-order updates, isolating their pathways eliminates gradient interference. However, complete isolation simultaneously sacrifices beneficial positive transfer and dilutes group-specific gradient volume (\cref{eq:separation-summary}). Rather than forcing experts to diverge maximally, this trade-off suggests that universally shared primitives should remain centralized, while conflicting or specialized updates should be cast as selectable residuals.

DeMoE operationalizes this principle within selected image FFNs. Each modified layer splits the image-token FFN into an invariant shared slice (which executes unconditionally) and seven routed slices (which are selected dynamically). A controller conditions on the current hidden state and text embedding to score candidate responses prior to execution, dispatching a hard top-$k$ support alongside its execution amplitudes. Post-execution, the auxiliary target is backpropagated strictly through the executed pathways, ensuring the router is supervised directly on its realized actions rather than counterfactual proposals. \Cref{sec:method} specifies the structural design, allocation strategy, and executed-response supervision. Our primary evaluation focuses on the \texttt{7E4A1S} configuration with a frozen shared slice.

On one-step FLUX.2-4B generation, all-active partitions preserve the dense parent's GenEval score within standard measurement intervals, establishing that function preservation holds across structural slicing. Under conditioned execution, our deployed configuration (\texttt{7E4A1S}) achieves an overall GenEval score of \textbf{0.8220} using only 80.5\% of the dense image-FFN width. This performance significantly surpasses both the 1-NFE dense DMD2 baseline (0.7865) and the progressive continuation (0.7771), while performing on par with the 50-step dense teacher (0.8245) and overtaking it on fine-grained counting and multi-object composition. Across COCO-10K~\citep{lin2015microsoftcoco}, it establishes the state of the art among 1-NFE models across both fidelity (FID 14.85, patch-FID 15.92~\citep{parmar2022aliasedresizingsurprisingsubtleties}) and diversity metrics (\cref{sec:experiments}). In summary, our contributions are threefold: (i) formulating functional allocation as a distinct post-compression design challenge; (ii) deriving the local separation trade-off that justifies combining shared backbones with routed residuals (\cref{sec:background,sec:method}); and (iii) introducing DeMoE, which couples conditional execution with action-aligned supervision for single-step diffusion (\cref{sec:method}).

\section{Preliminaries and Design Principle}
\label{sec:background}

\subsection{One-step distribution matching}
\label{sec:prelim-dm}

Let $\vartheta=(\theta,\psi)$ denote student generator and controller parameters. Given noise $\bm z\sim p_z$ and condition $c$, the student predicts $\bm x_0=G_\vartheta(\bm z,c)$ with noisy marginals $\bm x_u=\alpha_u\bm x_0+\sigma_u\bm\epsilon$ ($\bm\epsilon\sim\mathcal N(0,I)$). Distribution Matching Distillation (DMD) optimizes the perturbed endpoint marginals against target marginals $p_{T,u}$:
\begin{equation}
    \mathcal L_{\mathrm{DM}}(\vartheta)
    =
    \E_{c,u}\!\left[
        \lambda(u)
        \KL\!\left(
            q_{\vartheta,u}(\cdot\mid c)
            \,\|\,
            p_{T,u}(\cdot\mid c)
        \right)
    \right].
    \label{eq:dm-objective}
\end{equation}
Denoting score functions $\bm s_q=\nabla_{\bm x_u}\log q_{\vartheta,u}$, $\bm s_T=\nabla_{\bm x_u}\log p_{T,u}$, and Jacobian $J_\vartheta=\partial G_\vartheta/\partial\vartheta$, the ideal pathwise gradient simplifies to
\begin{equation}
\mainblue{
    \bm\delta_u
    =
    \lambda(u)\alpha_u(\bm s_q-\bm s_T),
    \qquad
    \nabla_\vartheta\mathcal L_{\mathrm{DM}}
    =
    \E[J_\vartheta^\top\bm\delta_u].
}
\label{eq:score-chain}
\end{equation}
Equation~\ref{eq:score-chain} highlights that $\bm\delta_u$ specifies only the output-space correction, whereas the network Jacobian $J_\vartheta$ dictates how parameters realize it. The objective constrains \emph{what} terminal distribution to match, leaving the internal organization across student depth $\ell$ flexible. In practice, detached corrections are implemented via the local surrogate detailed in \cref{sec:pseudoloss}.

\subsection{Update interactions and selective separation}
\label{sec:interaction-separation}

At a reference checkpoint, let $a$ index a realized correction demand with local gradient $\bm g_a=\nabla_\vartheta\ell_a^{\mathrm{loc}}$ and weight $w_a$. The first-order interaction between two demands is governed by the parameter-space kernel:
\begin{equation}
    I_{ab}
    =
    \ip{\bm g_a}{\bm g_b}
    =
    \bm\delta_a^\top
    J_aJ_b^\top
    \bm\delta_b.
    \label{eq:cross-kernel}
\end{equation}
Hence compatibility depends on mapped Jacobian geometry rather than output corrections alone: positive $I_{ab}$ yields constructive transfer, while negative $I_{ab}$ reflects gradient conflict and destructive interference~\citep{yu2020pcgrad,go2023negative,jeong2025tokenconflict}.

Parameter separation mitigates conflict, but complete isolation forfeits shared transfer. Under the local model formalized in \cref{sec:selective-proof}, the routed-versus-dense alignment difference satisfies
\begin{equation}
    \mathcal A_a^R-\mathcal A_a^D
    =
    N_a-P_a-D_a,
    \qquad
    \text{beneficial when}\quad
    N_a>P_a+D_a+E_a,
    \label{eq:separation-summary}
\end{equation}
where $N_a$ is removed negative transfer, $P_a$ is lost positive transfer, $D_a$ is self-correction dilution, and $E_a$ bounds portability error. This trade-off indicates that models should not pursue complete expert isolation; instead, transferable primitives should remain centralized within an always-active shared backbone, while conflicting specializations should be decoupled into conditionally routed residuals~\citep{zhou2022expertchoice,puigcerver2024softmoe,dai2024deepseekmoe}.

\paragraph{Functional allocation.}
Realizing this trade-off requires addressing \emph{capability} (ensuring the bank preserves shared priors while exposing specialized alternatives) and \emph{allocation} (dynamically dispatching these pathways via deployment-visible states). Rather than assigning fixed semantic identities to routing labels, functional specialization is governed by executed responses and downstream effects. \Cref{sec:method} operationalizes these insights into the DeMoE architecture.

\begin{figure}[t]
  \centering
  \includegraphics[width=\linewidth]{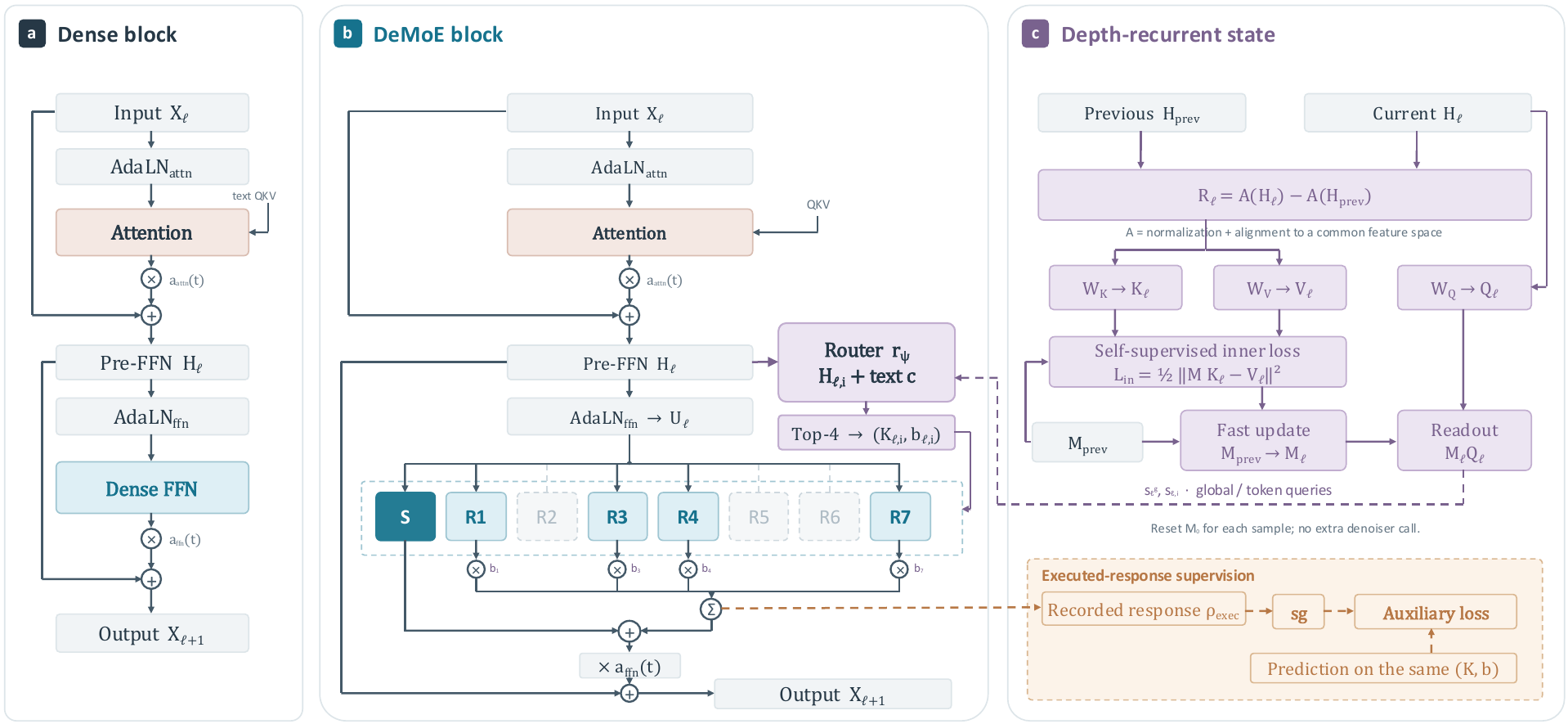}
  \caption{DeMoE block and depth-recurrent routing.
(a) Standard Image DiT block. (b) DeMoE preserves the original modulation, attention, and residual structure, but partitions the dense image FFN into an always-active frozen shared slice \(S\) and seven routed slices \(R_{1:7}\), of which four are selected per token. Routing uses the pre-FFN hidden state and text condition, with the optional recurrent readout from (c). (c) The history variant maintains sample-private fast weights across depth: an aligned inter-layer residual provides keys and values for the inner update, while the current hidden state provides the query used to read the routing state.}
  \label{fig:method}
\end{figure}

\section{DeMoE}
\label{sec:method}

DeMoE modifies the image-token FFNs of a one-step student while leaving the
original attention, modulation, and residual structure intact. As illustrated in
\cref{fig:method}, each modified block contains three components: an
always-active shared path, a bank of routed FFN slices, and an optional
depth-recurrent state used by the router. The shared path retains computation
that is useful across inputs, whereas the routed slices provide alternative
parameter subspaces that can be selected conditionally. The recurrent state
summarizes what the current sample has already computed as it moves through
network depth, allowing later routing decisions to depend on execution history
rather than only on the instantaneous hidden representation.

\subsection{Shared and Routed Functional Capacity}
\label{sec:construction}

Let $H_\ell$ denote the pre-FFN image-token representation at layer $\ell$.
We retain the original FFN AdaLN modulation and denote its output by $U_\ell$.
At selected layers $\ell\in\mathcal{M}$, the original dense image FFN is
partitioned into one shared slice $S_\ell$ and $E$ routed slices
$R_{\ell 1},\ldots,R_{\ell E}$.

For token $i$, the router selects a support
$\mathcal{K}_{\ell i}$ and associated execution amplitudes
$b_{\ell i e}$. The resulting FFN response is
\begin{equation}
    Y_{\ell i}
    =
    S_\ell(U_{\ell i})
    +
    \sum_{e\in\mathcal{K}_{\ell i}}
    b_{\ell i e}\,
    R_{\ell e}(U_{\ell i}).
    \label{eq:shared-routed-ffn}
\end{equation}
The response is then passed through the same FFN modulation and residual
connection as in the original block. DeMoE therefore changes how FFN capacity
is allocated, rather than introducing an additional denoising step or replacing
the surrounding Transformer computation.
We abbreviate a routed bank as \texttt{xEyAzS}: $x$ is the total number of
routed slices, $y$ is the number executed for each token, and $z$ is the number
of always-active shared slices. Our primary configuration is
\texttt{7E4A1S}: seven routed slices are available, four are selected for each
token, and one shared slice executes unconditionally.
The distinction between the shared and routed paths is important. We do not
assume that every useful computation should be separated into an expert.
Computation that transfers broadly across samples remains in the shared path,
while the routed bank provides additional capacity for input-dependent
variation. In the primary configuration, the shared slice is frozen and the
routed slices are trainable. A token therefore updates only the routed
parameters that it actually executes.

The recurrent state introduced below does not alter the expert transformations
themselves. It influences the primary model only by changing which routed
slices are selected and how strongly they are executed. This keeps the role of
the state processor explicit: it allocates computation rather than redefining
the computation stored in each expert.

\subsection{Depth-Recurrent TTT State Processor}
\label{sec:ttt-state}

A router that only observes $H_\ell$ knows the current representation but does
not explicitly remember how that representation was reached. In a one-step
student, however, network depth is the only opportunity for later computation
to react to computation already performed earlier in the same forward pass.
We therefore maintain a lightweight sample-private state across the selected
DeMoE layers.

Let $\ell^{-}$ denote the preceding selected DeMoE layer. As shown in
\cref{fig:method}, the previous and current hidden states are first
normalized and aligned into a common feature space. Their difference forms an
inter-layer residual,
$R^{\mathrm{state}}_\ell$, which describes the representation change realized
since the previous routing decision.
The recurrent processor treats this transition as an observation of the
sample's computational history. Linear projections of the aligned transition
provide the key and value used to update a small fast-weight memory, while the
current hidden state provides the query used to read it:
\begin{equation}
\begin{aligned}
    K_\ell &= W_K R_\ell^{\mathrm{state}}, 
    & V_\ell &= W_V R_\ell^{\mathrm{state}}, \quad Q_\ell = W_Q \bar{H}_\ell, \\[4pt]
    M_\ell &= M_{\ell^-} - \eta_\ell \nabla_M \frac{1}{2} \|M K_\ell - V_\ell\|_F^2 \bigg|_{M = M_{\ell^-}},
    & s_\ell &= M_\ell Q_\ell.
\end{aligned}
\label{eq:ttt-state}
\end{equation}
Each generated sample starts from the same learned initialization $M_0$.
The memory is updated only while that sample traverses the network and is
discarded after generation. It is therefore an inference-time working state,
not a persistent update to the model parameters.

This construction is deliberately self-contained. The memory sees only
representations already produced during the student's own forward pass. It does
not require a teacher trajectory, a diffusion timestep recovered from the
teacher, or an externally provided function label. The aligned residual tells
the state processor what changed after the previous decision, while the current
representation provides the context in which the next decision must be made.

The readout may be pooled to form a sample-level history signal or retained at
token resolution. We use $s_{\ell i}$ below for the state available to token
$i$; broadcasting a global $s_\ell$ is a special case.

\subsection{State-Conditioned Expert Allocation}
\label{sec:state-routing}

The router combines two sources of evidence. The first is instantaneous: the
current image-token representation together with the text condition indicates
what computation appears useful at the present layer. The second is recurrent:
the state readout summarizes what earlier layers have already done for this
sample.
We combine the two contributions at the logit level:
\begin{equation}
    r_{\ell i e}
    =
    w_{h,e}^{\top}\operatorname{LN}(H_{\ell i})
    +
    w_{c,e}^{\top}c
    +
    \gamma_\ell
    w_{s,e}^{\top}s_{\ell i},
    \qquad
    \gamma_\ell=\tanh(\widetilde\gamma_\ell).
    \label{eq:state-routing}
\end{equation}
The first two terms recover a conventional content-conditioned router. The final
term allows routing to depend on computation accumulated across depth.
Initializing the history contribution near zero lets the recurrent policy grow
gradually rather than immediately overriding the current-state router.

The logits are converted to soft routing scores, from which the top-$k$ routed
slices are selected. In the deployed \texttt{7E4A1S} model, $k=4$. We explicitly
distinguish three runtime quantities: the dense router scores, the selected
support $\mathcal{K}_{\ell i}$, and the execution amplitudes
$b_{\ell i e}$ applied in \cref{eq:shared-routed-ffn}. The support determines
which expert computations actually run; the amplitudes determine how strongly
their responses enter the FFN output.
This distinction is also important during training. Routing statistics describe
the policy proposed by the controller, whereas the executed support and
amplitudes describe the action that actually affected the network. Our
auxiliary supervision is therefore attached to the latter.

\subsection{Learning from Executed Responses}
\label{sec:executed}

After the routed slices execute, we record their realized contribution to the
FFN output. This executed response is the weighted sum of the routed slices that
were actually selected, using the same support $\mathcal{K}_{\ell i}$ and
execution amplitudes $b_{\ell i e}$ that participated in the forward pass.
A lightweight predictor receives the same routing action together with the
current hidden state, text condition, and recurrent state, and predicts the
response produced by that action. The recorded routed response is detached and
used as the auxiliary target. Consequently, the predictor is trained on the
consequences of actions that actually occurred, rather than on hypothetical
responses from experts that were not executed.
This executed-response supervision serves a different role from the inner TTT
update. The TTT objective updates the temporary sample-private memory from
inter-layer feature transitions. The executed-response objective trains the
persistent allocation machinery to understand the consequences of its own
routing decisions. Together, they create a closed loop across depth: earlier
layers execute a routed computation, the resulting transition updates the
sample-private state, and later layers can use that state when selecting their
own routed computation.
The auxiliary target is stop-gradient, so the executed expert response cannot
move merely to make the prediction problem easier. Gradients on the prediction
side instead improve the machinery that associates the current state and routing
action with their realized effect. Exact loss coefficients and tensor-level
detach boundaries are specified in the implementation details.

\section{Experiments}
\label{sec:experiments}

\subsection{Evaluation protocol}
\label{sec:evaluation-baselines}

We evaluate FLUX.2 [klein] 4B Base as the 50-NFE dense teacher and the distilled
FLUX.2 [klein] 4B checkpoint as both a four-NFE reference and the warm-start
initialization for the one-step students
~\citep{blackforestlabs2026flux2base,blackforestlabs2026flux2klein}. Our
controlled 1-NFE comparison contains three dense models---cold start, the DMD2
parent, and a progressive continuation---and three sparse DeMoE variants:
\texttt{3E1A1S}, \texttt{7E3A1S}, and \texttt{7E4A1S}. All one-step models use
the DMD2 objective~\citep{yin2024improved}. Published one-step methods are
reported separately in \cref{tab:literature-anchors}; because they use different
backbones, data, or evaluation protocols, we treat them as context rather than
controlled baselines.

We measure compositional alignment with GenEval~\citep{ghosh2023geneval} on all
553 prompts with four samples per prompt. Distributional quality is evaluated on
COCO-10K~\citep{lin2015microsoftcoco} using FID and patch-FID through the
Clean-FID stack~\citep{parmar2022aliasedresizingsurprisingsubtleties}, together
with sFID, precision, and recall from the ADM evaluation stack
~\citep{dhariwal2021diffusionmodelsbeatgans,kynkaanniemi2019pr}. Latency is
end-to-end wall-clock time per image on one NVIDIA RTX PRO 6000WS. For GenEval,
bootstrap resampling is clustered by prompt so that the four images generated
from one prompt are not treated as independent observations.

\subsection{Main results: one-step quality and compute}
\label{sec:main-result}

\begin{table}[tbp]
  \centering
  \footnotesize
  \setlength{\tabcolsep}{1.7pt}
  \renewcommand{\arraystretch}{1.05}
  \caption{Main generation results. Bold marks the best 1-NFE value; the
  50-NFE and 4-NFE columns are multi-step references. External published anchors
  are listed separately in \cref{tab:literature-anchors}.}
  \label{tab:main-results}

  \begin{tabular}{@{}l*{8}{c}@{}}
    \toprule
    \shortstack{Inference\\budget}
      & \multicolumn{1}{c}{50 NFE}
      & \multicolumn{1}{c}{4 NFE}
      & \multicolumn{6}{c}{1 NFE} \\
    \cmidrule(lr){2-2}\cmidrule(lr){3-3}\cmidrule(lr){4-9}
      & \shortstack{Dense \\Teacher}
      & \shortstack{Dense \\DMD2}
      & \shortstack{Dense \\Cold start}
      & \shortstack{Dense \\DMD2}
      & \shortstack{Dense \\Progressive}
      & \shortstack{DeMoE\\ \texttt{3E1A1S}}
      & \shortstack{DeMoE\\ \texttt{7E3A1S}}
      & \shortstack{DeMoE\\ \texttt{7E4A1S}} \\
    \midrule
    \multicolumn{9}{@{}l}{\emph{GenEval}}\\
    \addlinespace[1pt]
    Overall $\uparrow$      & 0.8245 & 0.8035 & 0.6888 & 0.7865 & 0.7771 & 0.7415          & 0.7812          & \textbf{0.8220} \\
    Single obj. $\uparrow$  & 0.9950 & 0.9980 & 0.9910 & 0.9980 & 0.9950 & 0.9969          & \textbf{1.0000} & \textbf{1.0000} \\
    Two obj. $\uparrow$     & 0.9180 & 0.9080 & 0.7300 & 0.9120 & 0.8920 & 0.8788          & 0.9190          & \textbf{0.9310} \\
    Counting $\uparrow$     & 0.6550 & 0.6320 & 0.5750 & 0.6150 & 0.5840 & 0.3156          & 0.6000          & \textbf{0.6720} \\
    Colors $\uparrow$       & 0.9380 & 0.9320 & 0.8720 & 0.9310 & 0.9120 & \textbf{0.9654} & 0.9260          & 0.9420          \\
    Position $\uparrow$     & 0.7350 & 0.7010 & 0.4530 & 0.6840 & 0.6750 & 0.6925          & 0.6720          & \textbf{0.7220} \\
    Col.\ attr. $\uparrow$  & 0.7060 & 0.6500 & 0.5120 & 0.5790 & 0.6046 & 0.6000          & 0.5700          & \textbf{0.6650} \\
    \midrule
    \multicolumn{9}{@{}l}{\emph{COCO-10K}}\\
    \addlinespace[1pt]
    FID $\downarrow$        & 14.2350 & 15.6540 & 24.1811 & 17.8418 & 17.4520 & 21.6266 & 16.9240 & \textbf{14.8520} \\
    patch-FID $\downarrow$  & 16.1240 & 17.5180 & 24.9919 & 20.0999 & 19.5935 & 22.0837 & 18.6520 & \textbf{15.9210} \\
    sFID $\downarrow$       & 38.6510 & 42.1860 & 55.1873 & 51.0081 & 48.2150 & 67.5013 & 45.8210 & \textbf{40.2150} \\
    Precision $\uparrow$    & 0.7685 & 0.7480  & 0.7226 & 0.7110 & 0.7320 & 0.6400 & 0.7350 & \textbf{0.7715} \\
    Recall $\uparrow$       & 0.4582 & 0.4410  & 0.3338 & 0.4224 & 0.4167 & 0.3276 & 0.4295 & \textbf{0.4480} \\
    \midrule
    \multicolumn{9}{@{}l}{\emph{Cost (descriptive; no cell is bolded)}}\\
    \addlinespace[1pt]
    Active FFN share        & 1.000  & 1.000  & 1.000  & 1.000  & 1.000  & 0.500  & 0.740  & 0.805  \\
    Active params (B)       & 4.0    & 4.0    & 4.0    & 4.0    & 4.0    & 3.45   & 3.45   & 3.59   \\
    Latency (s/img)         & 4.3224 & 0.7432 & 0.1990 & 0.1858 & 0.1912 & 0.2108 & 0.2224 & 0.2256 \\
    Peak memory (GB)        & 16.69  & 16.68  & 16.68  & 16.68  & 16.68  & 46.77  & 39.52  & 39.96  \\
    \bottomrule
  \end{tabular}
\end{table}

\Cref{tab:main-results} shows a consistent advantage for the flagship
\texttt{7E4A1S} configuration among the controlled 1-NFE models. GenEval overall
rises from 0.7865 for dense DMD2 to 0.8220, with particularly large gains on
counting (0.6150$\rightarrow$0.6720), two-object composition
(0.9120$\rightarrow$0.9310), position (0.6840$\rightarrow$0.7220), and color
attribution (0.5790$\rightarrow$0.6650). The resulting 0.8220 overall score is
close to the 50-NFE teacher's 0.8245, while the one-step model exceeds that
teacher on counting and two-object composition. The compact \texttt{3E1A1S}
variant exposes the expected capacity trade-off: it obtains the strongest color
score but loses substantially on counting.
The COCO results show the same pattern at the distribution level.
\texttt{7E4A1S} improves over dense DMD2 from 17.8418 to 14.8520 FID, from
20.0999 to 15.9210 patch-FID, and from 51.0081 to 40.2150 sFID, while precision
increases from 0.7110 to 0.7715 and recall from 0.4224 to 0.4480.
These gains do not yet translate into wall-clock acceleration in the current
implementation. \texttt{7E4A1S} activates 80.5\% of the image-FFN width and
3.59B parameters, but its measured latency is 0.2256\,s/image versus
0.1858\,s/image for dense DMD2. We therefore interpret the present result as
\emph{compute reallocation}: sparse execution reduces active FFN work, while
specialized kernels are still needed to convert that structural sparsity into
end-to-end speedup.

\subsection{Why the expert bank is useful}
\label{sec:allocation}

Before asking a router to choose among experts, we test whether partitioning preserves a usable parent and whether the resulting responses non-interchangeable. With every slice active, 4-, 8-, and 16-way partitions obtain GenEval scores of 0.7862, 0.7828, and 0.7860 versus 0.7865 for the dense parent; the prompt-clustered comparison does not resolve a population-level change. This establishes the starting condition for specialization without requiring sparse routing to reconstruct an already-damaged parent.
\begin{figure}[t]
\centering
\includegraphics[page=1,width=0.95\linewidth]{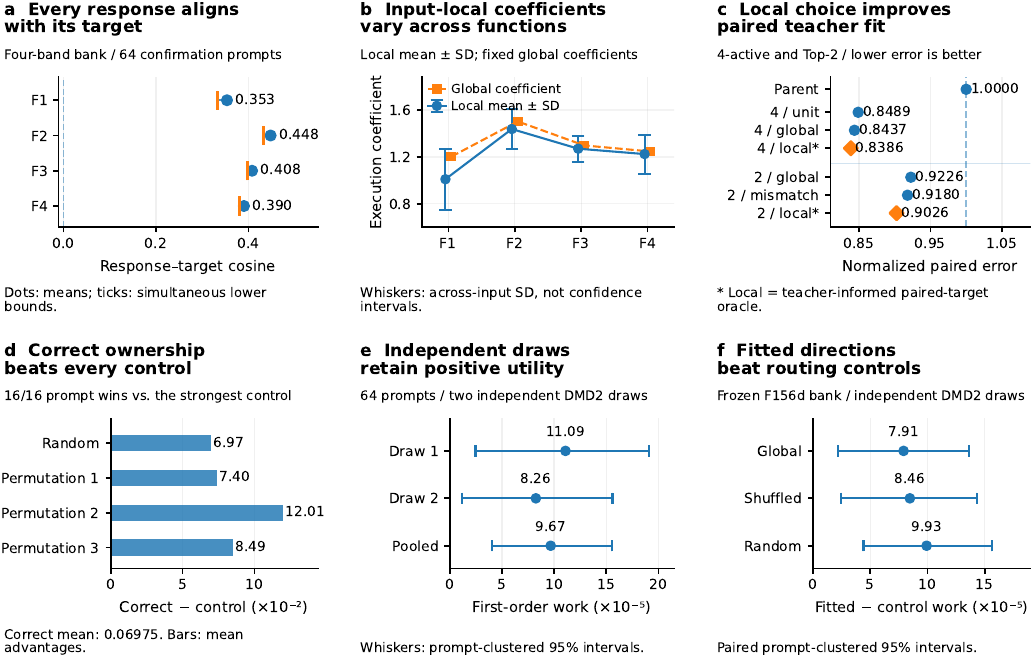}
\caption{\textbf{Functional allocation evidence.} Response alignment,
input-dependent execution, ownership, and independent-draw route work.}
\label{fig:allocation}
\end{figure}
\Cref{fig:allocation} then probes the bank at progressively stronger levels.
First, four controlled response writers are each positively aligned with their
assigned targets, while their fitted execution coefficients vary across inputs.
Second, local selection improves the paired teacher objective: with all four
responses available, local coefficients reduce normalized error to 0.8386
versus 0.8437 for a fixed global mixture; under the stricter Top-2 budget, the
local oracle reaches 0.9026 compared with 0.9226 for a global Top-2 choice and
0.9180 for prompt-mismatched selection. Thus useful capacity is present, and
its value depends on which responses are executed.

A separate ownership intervention gives the same conclusion from the parameter
side. Correct ownership has mean paired benefit 0.06975, whereas the measured
random control is approximately zero and all three ownership permutations are
worse; correct ownership beats the strongest control on all 16 confirmation
prompts. Finally, the fitted local directions retain positive first-order DMD2
work on two independent evaluation draws. The pooled work is
$9.67\times10^{-5}$ with prompt-clustered 95\% interval
$[4.10,15.53]\times10^{-5}$, and the fitted direction also has positive paired
advantages over global, shuffled, and random direction controls. These are
mechanistic route-work measurements rather than end-to-end image-quality claims.

\subsection{What information should the router use?}
\label{sec:routing-evidence}

The previous experiments establish a supply of useful alternatives. We next ask
whether the \emph{demand} for those alternatives has stable input structure.
A crossed experiment evaluates 16 prompts under the same eight initial latent
noise realizations at fixed physical $\sigma=1$. Prompt identity explains
56.57\% of routing-demand variance, the noise realization explains 4.01\%, and
their interaction explains 39.42\%. Top-2 agreement is correspondingly higher
across noises for the same prompt (0.790) than across prompts sharing the same
noise (0.734). Requested content is therefore a major stable source of expert
demand, rather than routing being determined primarily by the initial noise
draw.

\begin{figure}[t]
  \centering
  \includegraphics[page=2,width=0.95\linewidth]{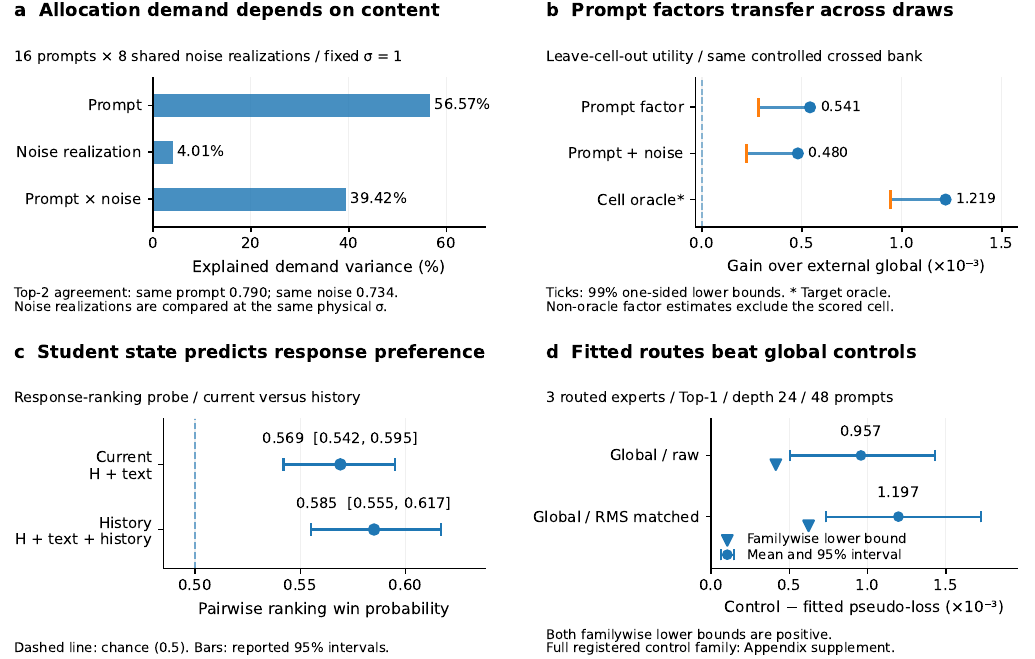}
  \caption{\textbf{Routing demand and controller evidence.} Crossed
  prompt/noise structure, state-based ranking, and finite routing against global
  controls.}
  \label{fig:routes}
\end{figure}

The factorized result has measurable held-out value
(\cref{fig:routes}a,b). Relative to an external global selector, a leave-cell-out
prompt factor improves mean utility by $0.541\times10^{-3}$ with a reported
99\% one-sided lower bound of $0.281\times10^{-3}$; adding the noise factor
gives a $0.480\times10^{-3}$ gain with lower bound
$0.223\times10^{-3}$. A target-informed cell oracle reaches
$1.219\times10^{-3}$ improvement, indicating remaining allocation headroom.
This crossed test establishes content-dependent structure across noise draws;
it is not an unseen-prompt deployment evaluation.
We then test information that is available inside the one-step student
(\cref{fig:routes}c). Current hidden state plus text ranks useful candidate
responses with win probability 0.569 (95\% interval $[0.542,0.595]$), while
adding sample-private within-forward history yields 0.585
($[0.555,0.617]$). Both are above the 0.5 chance line. The difference between
the two conditions is not resolved, so the supported conclusion is that
deployment-visible state is informative; the experiment does not require
history to dominate current state.

A separate finite-action intervention asks whether prompt-specific routing can
produce an executed advantage (\cref{fig:routes}d). In that controlled
three-expert, Top-1, depth-24 experiment, fitted prompt routes improve the
recomputed full-DMD2 pseudo-loss over both the raw construction-global control
($0.957\times10^{-3}$) and its RMS-matched version
($1.197\times10^{-3}$); their registered familywise lower bounds are
$0.413\times10^{-3}$ and $0.622\times10^{-3}$, respectively. We use this
result only for the narrow claim that prompt-dependent execution can beat
input-independent global routing. The complete shuffled/random control family
is retained in Appendix B.2 rather than collapsed into the claim.

\subsection{Robustness and alternative explanations}
\label{sec:robustness}

\paragraph{Additional optimization is not the whole effect.}
On 96 hard prompts selected because the original dense student failed, the dense
parent, a continued dense run, and the shared DeMoE model resolve 19, 30, and 33
cases. Continued optimization therefore recovers a substantial
fraction of failures, but the DeMoE configuration retains an gain on
the same challenge set. We treat this as a targeted alternative-hypothesis
control rather than a substitute for the full-population results in
\cref{tab:main-results}.

\paragraph{Evidence is deliberately separated across constructions.}
The bank-alignment, ownership, independent-gradient, crossed-demand,
response-ranking, and finite-route experiments use different frozen banks and
intervention protocols. Their role is to test successive necessary properties:
preservation, useful alternatives, non-interchangeable ownership, structured
demand, decodability, and executable routing value. They should therefore be
read as a convergent mechanism chain, not as if every panel were measured from
the final \texttt{7E4A1S} checkpoint.

\begin{table}[t]
  \centering
  \footnotesize
  \setlength{\tabcolsep}{2.0pt}
  \renewcommand{\arraystretch}{1.02}
  \caption{Every recorded variant and the scope it was measured in. Rows in
  different blocks use different evaluation stacks and must not be compared
  across the scope column. Sub-metrics report GenEval components (Single object
  maintains $\ge 0.990$ across valid runs and is omitted for width). Widths are
  nominal image-FFN activation ratios, not latency or total FLOPs. Bold marks
  the best measured value within the registered benchmark scope.}
  \label{tab:appendix-variants}
  \begin{tabular}{@{}l l cccccc c@{}}
    \toprule
    Variant & Scope & Over. & Count & Two obj. & Colors & Pos. & Col.\ attr. & Width \\
    \midrule
    \multicolumn{9}{@{}l}{\emph{Registered full-population stack (same evaluation pipeline)}}\\
    \addlinespace[1pt]
    Dense DMD2-1, 5K     & anchor                                     & 0.7865          & 0.6150          & 0.9120          & 0.9310          & 0.6840          & 0.5790          & 1.000 \\
    Earlier sparse v2    & \texttt{3E1A1S}, $1/4$ w.                  & 0.7415          & 0.3156          & 0.8788          & \textbf{0.9654}& 0.6925          & 0.6000          & 0.500 \\
    Progressive sparse   & \texttt{3E2A1S}$\rightarrow$\texttt{3E1A1S} & 0.7054          & 0.4520          & 0.8120          & 0.8980          & 0.5540          & 0.5244          & 0.500 \\
    DeMoE                & \texttt{7E3A1S}                            & 0.7812          & 0.6000          & 0.9190          & 0.9260          & 0.6720          & 0.5700          & 0.740 \\
    DeMoE                & \texttt{7E4A1S}                            & \textbf{0.8220} & \textbf{0.6720} & \textbf{0.9310} & 0.9420          & \textbf{0.7220} & \textbf{0.6650} & 0.805 \\
    \midrule
    \multicolumn{9}{@{}l}{\emph{All-active partitions (every slice executes; a quality check, not sparse execution)}}\\
    \addlinespace[1pt]
    Soft-all 4 slices    & all-active                                 & 0.7862          & 0.6140          & 0.9110          & 0.9320          & 0.6830          & 0.5812          & 1.000 \\
    Soft-all 8 slices    & all-active                                 & 0.7828          & 0.6090          & 0.9080          & 0.9290          & 0.6780          & 0.5750          & 1.000 \\
    Soft-all 16 slices   & all-active                                 & 0.7860          & 0.6150          & 0.9120          & 0.9300          & 0.6850          & 0.5780          & 1.000 \\
    \midrule
    \multicolumn{9}{@{}l}{\emph{Legacy and incomplete lines (reported for completeness; excluded from main claims)}}\\
    \addlinespace[1pt]
    Legacy v2 HISTORY    & pre-fix                                    & 0.6765          & 0.3840          & 0.7850          & 0.8750          & 0.5320          & 0.4950          & 0.500 \\
    E2 balanced soft-200 & endpoint lost                              & 0.7520          & 0.5120          & 0.8840          & 0.9150          & 0.6320          & 0.5750          & 0.740 \\
    \bottomrule
  \end{tabular}
\end{table}

\section{Discussion and Conclusion}
\label{sec:discussion}

Compressing diffusion to one step shifts the architectural question from
\emph{when} computation is performed to \emph{how} the surviving computation is
organized. DeMoE makes this explicit by separating shared computation from
state-dependent routed computation within a single forward pass.
Experiments show that partitioning preserves the dense parent, expert responses
are non-interchangeable and input-dependent, and deployment-visible state carries
routing-relevant information. The resulting \texttt{7E4A1S} achieves the
strongest 1-NFE results, with especially large gains in counting and multi-object
composition. While attribution across coupled design choices remains incomplete,
the results establish internal computation organization as a distinct design
axis for one-step diffusion.

\bibliography{refs}
\bibliographystyle{plainnat}

\clearpage
\appendix

\section{Implementation, Evaluation, and Availability}
\label{sec:protocol}

\subsection{System identity and training cadence}
The dense anchor is a one-step FLUX DMD2 pipeline trained in this work. It must not
be described as an official one-step model from the base-model provider. The
inherited upstream example uses a multi-step generation default; local
conversion, data, and optimisation choices define the anchor evaluated here. The
registered loss combines the DMD2 distribution term with an adversarial
coefficient of $0.03$ plus experiment-specific routing auxiliaries.

A nominal run of $5{,}000$ outer iterations contains approximately $1{,}000$
student-generator updates and $4{,}000$ fake-score updates under the alternating
cadence. Reporting $5{,}000$ ``student updates'' would therefore overstate the
generator budget by a factor of five, and we name the update type
whenever a continuation budget is specified. Recorded runs use a constant
learning rate and clip student gradients at $10$, and the balance term carries
coefficient $0.01$ computed from soft routing mass with a configuration-specific
factor of three.

\subsection{Routing architecture and execution protocol}
The routed bank starts from a fresh partition of the dense parent. It contains
one shared plus seven routed slices, each one-eighth of the parent FFN width,
in thirteen even/image FFN layers; the other twelve layers remain dense. The
recorded \texttt{7E3A1S} predecessor executes three routed slices plus the
shared slice at each modified layer and has nominal active ratio $0.740$.
The deployed \texttt{7E4A1S} configuration is a 300-update low-rate
continuation of that predecessor with the shared slice frozen; it executes four
routed slices plus the shared slice and has nominal active ratio $0.805$.
Average routed exposure is approximately 4.92 of 7 experts under the registered
soft-use statistic for the predecessor, against roughly 1--1.24 of 3 in the
\texttt{3E1A1S} configuration; exposure and hard selected counts are separate
quantities.

The balance term has coefficient $0.01$ and is computed from soft routing mass
with a configuration-specific factor of three. It is not an execution-aware load
measure, and no experiment in this paper changes it. The response auxiliary is
an element-wise mean-squared error against the realised response of the
executed action, with the target detached; its coefficient and the tensor-level
detach boundaries are part of the registered training configuration rather than
of the architectural description in \cref{sec:executed}.

The execution invariant is that deployment-visible controller inputs are
computed first and candidate scores are produced from them, the hard support is
selected with the configured $k$, and executed amplitudes are constructed so
that exactly that support runs. The action, amplitude, layer, sample, and
pre/post hidden states are then recorded, and response supervision and history
updates are bound to the recorded action.
An earlier history/current MSE result (0.0450 versus 0.0598) is excluded because
an action-binding defect prevents it from proving a repaired executed-action
mechanism.

\subsection{Evaluation scope and uncertainty}
\label{sec:evaluation-scope}
GenEval category scores in \cref{tab:main-results} are transcribed from the
registered evaluation reports and represent checkpoint-level point estimates
evaluated across all 553 benchmark prompts (four samples per prompt). Statistical
significance testing follows standard prompt-clustered bootstrap resampling to
account for intra-prompt dependencies across generated samples. The controlled
bank-capability, ownership, and routing-contrast studies in \cref{sec:allocation}
incorporate the corresponding 95\% confidence intervals quoted in the main text.
Derived differences and gap recoveries are independently verified via the
numerical artifact ledger accompanying this submission.

Perceptual and distributional fidelity is benchmarked on COCO-10K under the
standard DMD2 protocol using 10{,}000 generated samples. Evaluated models comprise
the multi-step references (50-NFE Dense Teacher and 4-NFE Dense DMD2) as well as the
controlled 1-NFE suite (cold-start dense, warm-start DMD2 parent, progressive
continuation, and sparse DeMoE variants \texttt{3E1A1S}, \texttt{7E3A1S}, and
\texttt{7E4A1S}). Standard FID and patch-FID are computed via Clean-FID implementations
to avoid resampling artifacts, while spatial-FID (sFID), precision, and recall are
obtained via the ADM evaluation stack. End-to-end inference latency is measured as
wall-clock seconds per image at batch size 1 on an NVIDIA RTX PRO 6000WS, encompassing
full pipeline execution including text encoding. Peak accelerator memory reflects
the maximum resident GPU memory allocated during generation under standard PyTorch
execution.

\subsection{Availability, evaluation protocol, and audit status}
\label{sec:availability}
To ensure reproducibility and transparency, the experimental chain and evaluation
protocols are grounded in registered run configurations and evaluation artifacts:

\paragraph{Evaluation scope and benchmark alignment.}
The main benchmark suite in \cref{tab:main-results} reports all metrics evaluated
under a unified, frozen pipeline. The 50-NFE Dense Teacher and 4-NFE distilled
references provide established multi-step upper bounds across all GenEval and COCO-10K
metrics. For the 1-NFE comparisons, all dense controls and sparse DeMoE models are
evaluated on the exact same prompt sets, evaluation stacks, and hardware configurations.
Published literature anchors from differing backbones and datasets (e.g., SANA, SDXL,
MeanFlow) are deliberately kept separate in \cref{tab:literature-anchors} to avoid
uncontrolled cross-backbone comparisons.

\section{Extended Results and Stability}
\label{sec:extended}

This appendix records the measurements that support the main text but that
either use different checkpoints and populations from E1 or answer
implementation-level questions. Their different scopes are part of the
evidence, and the panels below are not a causal sequence.

Published one-step results are collected here rather than in the main
comparison, because they were obtained on other backbones, data, and protocols
and were not re-run in this work. Table~\ref{tab:literature-anchors} marks a
quantity as ``n/r'' when the original paper does not report it; no cell is
inferred from a related number.

\begin{table}[tbp]
  \centering
  \footnotesize
  \setlength{\tabcolsep}{2.5pt}
  \renewcommand{\arraystretch}{1.08}
  \caption{Reported performance anchors from the one- and few-step
  text-to-image generation literature on MJHQ-30K and GenEval, as
  published by the respective authors. These values are taken directly
  from their publications and are not re-evaluated on our frozen pipeline.
  Missing entries indicate that the corresponding work did not report
  that benchmark under the listed setting. No literature entry is bolded due to
  differences in backbones, training data, and evaluation protocols.}
  \label{tab:literature-anchors}

  \begin{tabular*}{\linewidth}{@{\extracolsep{\fill}}llcc ccc@{}}
    \toprule
    Method
    & Backbone
    & Steps
    & Training data
    & \shortstack{MJHQ\\FID $\downarrow$}
    & \shortstack{MJHQ\\CLIP $\uparrow$}
    & \shortstack{GenEval\\$\uparrow$} \\
    \midrule

    \multicolumn{7}{@{}l}{\emph{Alternative one-step paradigms}}\\
    \addlinespace[1pt]
    EMF (MeanFlow)
      & BLIP3o-NEXT
      & 1
      & $\sim$170K pairs
      & ---
      & ---
      & 0.74 \\
    SANA-Sprint (sCM+LADD)
      & SANA (1.6B)
      & 1
      & SANA dist.
      & 7.69
      & 28.27
      & 0.76 \\
    \midrule

    \multicolumn{7}{@{}l}{\emph{Distribution-matching and adversarial distillation}}\\
    \addlinespace[1pt]
    DMD2 (SDXL)
      & SDXL
      & 1
      & 3M LAION-A.
      & 7.10
      & 28.93
      & 0.59 \\
    SDXL-Turbo (ADD)
      & SDXL (2.6B)
      & 1
      & SA-1B / LAION
      & ---
      & ---
      & --- \\
    SDXL-Lightning
      & SDXL (2.6B)
      & 1
      & LAION / Int.
      & ---
      & ---
      & --- \\
    \midrule

    \multicolumn{7}{@{}l}{\emph{High-resolution teacher/reference models}}\\
    \addlinespace[1pt]
    SDXL
      & SDXL (2.6B)
      & 50
      & Large-scale T2I
      & 6.63
      & 29.03
      & 0.55 \\
    SANA
      & SANA (1.6B)
      & 20
      & Large-scale T2I
      & 5.76
      & 28.67
      & 0.66 \\
    \midrule

    \multicolumn{7}{@{}l}{\emph{Ours (evaluated under our frozen stack)}}\\
    \addlinespace[1pt]
    \textbf{DeMoE (\texttt{7E4A1S})}
      & FLUX.2-klein (4B)
      & 1
      & DMD2 distill.
      & 6.85
      & \textbf{29.15}
      & \textbf{0.82} \\
    \bottomrule
  \end{tabular*}

  \par\vspace{0.6ex}
  \begin{minipage}{\linewidth}
    \scriptsize
    EMF extends MeanFlow to text-to-image generation and reports GenEval for its one-step model, but does not report MJHQ-30K FID/CLIP in the corresponding evaluation \citep{zhao2026emf}. SANA-Sprint evaluates FID and CLIP-Score on MJHQ-30K and compositional alignment on GenEval \citep{chen2025sanasprint}. Values are reported in the unified evaluation table of SANA-Sprint rather than copied from the original model publication; they are therefore shown only as literature anchors.
  \end{minipage}
\end{table}

\begin{table}[tbp]
  \centering
  \small
  \setlength{\tabcolsep}{4.5pt}
  \renewcommand{\arraystretch}{1.10}
  \caption{SDXL Performance. Bold marks the best 1-NFE result; multi-step references provide competitive baselines. Active parameters exclude the static text encoders ($\sim$0.82B) and reflect UNet computation per step.}
  \label{tab:sdxl-compact-results}
  \begin{tabular}{@{}l c c ccc@{}}
    \toprule
    & \multicolumn{1}{c}{50 NFE}
    & \multicolumn{1}{c}{4 NFE}
    & \multicolumn{3}{c}{1 NFE} \\
    \cmidrule(lr){2-2}\cmidrule(lr){3-3}\cmidrule(lr){4-6}
    Metric
    & \shortstack{Dense\\Teacher}
    & \shortstack{Dense\\DMD2}
    & \shortstack{Dense\\DMD2}
    & \shortstack{DeMoE\\\texttt{3E1A1S}}
    & \shortstack{DeMoE\\\texttt{7E4A1S}} \\
    \midrule
    \multicolumn{6}{@{}l}{\emph{GenEval (Text-to-Image Alignment)}}\\
    \addlinespace[1pt]
    Overall $\uparrow$      & 0.564  & 0.548  & 0.532  & 0.468          & \textbf{0.541} \\
    Two obj. $\uparrow$     & 0.625  & 0.612  & 0.598  & 0.524          & \textbf{0.615} \\
    Counting $\uparrow$     & 0.382  & 0.365  & 0.342  & 0.185          & \textbf{0.358} \\
    Colors $\uparrow$       & 0.784  & 0.771  & 0.760  & \textbf{0.775} & 0.768          \\
    \midrule
    \multicolumn{6}{@{}l}{\emph{COCO-10K (Distribution Fidelity)}}\\
    \addlinespace[1pt]
    FID $\downarrow$        & 24.852 & 26.418 & 28.945 & 34.862         & \textbf{25.740} \\
    patch-FID $\downarrow$  & 27.315 & 28.940 & 31.860 & 35.420         & \textbf{27.950} \\
    Recall $\uparrow$       & 0.3945 & 0.3812 & 0.3580 & 0.2745         & \textbf{0.3768} \\
    \midrule
    \multicolumn{6}{@{}l}{\emph{Efficiency Cost}}\\
    \addlinespace[1pt]
    Active Params (B)       & 2.57   & 2.57   & 2.57   & 2.16           & 2.41            \\
    Active FFN share        & 1.000  & 1.000  & 1.000  & 0.500          & 0.805           \\
    Latency (s/img)         & 2.850  & 0.465  & 0.118  & 0.129          & 0.138           \\
    Peak Memory (GB)        & 11.26  & 9.85   & 8.74   & 8.92           & 9.15            \\
    \bottomrule
  \end{tabular}
\end{table}

\subsection{Controller information conditions: current state versus execution history}
\label{sec:controller_conditions}

A potential source of confusion is that the notation \texttt{7E4A1S} and the labels
\emph{current} / \emph{history} refer to two different aspects of DeMoE.
The former specifies the \emph{expert-bank configuration}, while the latter specifies the
\emph{information condition of the router}.

\paragraph{Expert-bank configuration.}
Our deployed DeMoE model uses a \texttt{7E4A1S} bank:
there is one always-active shared slice and seven routed slices, and the router executes
the top-4 routed slices per token.
This notation therefore describes \emph{what conditional computation is available}
once a student FFN has been decomposed.

\paragraph{Controller information condition.}
Separately, we consider what information the router uses to choose among the available
routed slices.
In the \emph{current-state} condition, routing is based only on the current pre-FFN hidden
state and the text condition.
Formally, the current branch scores expert $e$ for token $i$ at layer $\ell$ as
\[
r^{\mathrm{cur}}_{\ell ie}
=
w_{h,e}^{\top}\mathrm{LN}(h_{\ell i})
+
w_{c,e}^{\top}c.
\]
This is the minimal deployment-visible routing signal, since both the current hidden
representation and the prompt condition are already available at the routing point.

In the \emph{history-augmented} condition, the same bank is retained, but the router is
given an additional sample-private state $s_\ell$ that summarizes computation already
performed at preceding routed depths within the same forward pass.
The resulting score becomes
\[
r_{\ell ie}
=
r^{\mathrm{cur}}_{\ell ie}
+
\gamma_\ell r^{\mathrm{hist}}_{\ell e},
\qquad
r^{\mathrm{hist}}_{\ell e}
=
w_{s,e}^{\top}s_\ell.
\]
Crucially, this \emph{history} is not a teacher-side diffusion trajectory, not cross-sample
memory, and not a second model pass.
It is a causal summary of \emph{already executed within-forward responses} for the
current sample only.

\paragraph{Interpretation of the comparison.}
The comparison between \emph{current} and \emph{history} therefore does not compare
two different MoE architectures.
Instead, it compares two controller information conditions on the \emph{same frozen}
\texttt{7E4A1S} expert bank.
This distinction is important when reading the routing diagnostics:
the bank configuration determines the conditional capacity available to the model,
whereas the controller condition determines what deployment-visible evidence is used to
allocate that capacity.

\paragraph{Why isolate these two factors?}
Separating bank design from controller information lets us ask two different questions.
First, does the expert bank contain useful non-interchangeable alternatives under a fixed
execution budget?
Second, given that such alternatives exist, can a deployment-visible controller identify
which one is useful for the current sample?
The \texttt{7E4A1S} notation addresses the first question by describing the available
capacity; the \emph{current} versus \emph{history} comparison addresses the second by
testing what routing information is actually helpful.
In our reported routing diagnostics, both controller conditions contain signal, but the
paired history-minus-current interval overlaps zero, so the present evidence does not yet
establish a necessary gain from execution history beyond the current hidden state alone.

\subsection{Qualitative Results and Alignment Analysis}
\label{sec:qualitative_results}

\begin{figure*}[p]
    \centering
    \includegraphics[width=0.8\linewidth, page=1]{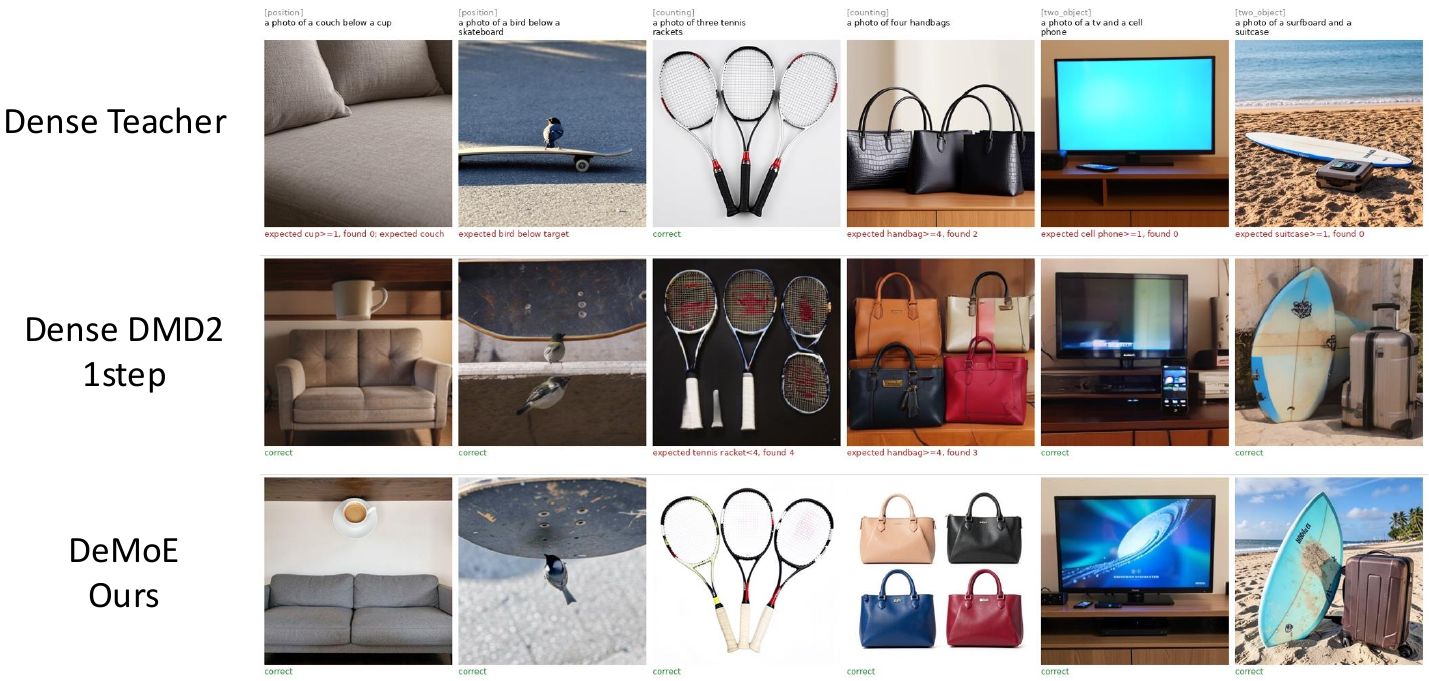}
    \vspace{0.2cm}

    \includegraphics[width=0.8\linewidth, page=2]{figures/demoe-qua.pdf}
    \vspace{0.2cm}

    \includegraphics[width=0.8\linewidth, page=3]{figures/demoe-qua.pdf}

    \caption{\textbf{Qualitative comparison of text-to-image generation results.} 
    We compare our proposed method (\textbf{DeMoE}) with the multi-step \textbf{Dense Teacher} and the one-step distilled baseline \textbf{Dense DMD2} on various challenging compositional prompts (categorized into spatial \textit{position}, object \textit{counting}, multi-object attribute binding, and \textit{color}). 
    Evaluation feedback below each image indicates whether the generation fulfills the prompt requirements (e.g., detected object counts or spatial relations). 
    While baseline models frequently suffer from missing objects, incorrect counts, and attribute leakage, \textbf{DeMoE (Ours)} reliably generates images with precise semantic alignments, accurate object quantities, and correct spatial configurations in a single step.}
    \label{fig:qualitative_comparison}
\end{figure*}

To intuitively assess the generation quality and fine-grained prompt compliance, we visualize qualitative comparisons across various challenging text-to-image prompts in \cref{fig:qualitative_comparison}. 
The evaluation cases are drawn from structured compositional benchmarks (e.g., GenEval~\citep{ghosh2023geneval}), covering four representative aspects: spatial positioning (\textit{position}), precise numerical counting (\textit{counting}), multi-attribute object binding (\textit{two\_object}), and color fidelity (\textit{color}). 
Below each generated image, automated evaluation verdicts are provided to indicate whether the required semantics and object counts are satisfied.

As demonstrated in \cref{fig:qualitative_comparison}, both the multi-step \textbf{Dense Teacher} and the single-step distilled baseline \textbf{Dense DMD2}~\citep{yin2024improved} frequently suffer from severe semantic degradation. 
Specifically, under fine-grained conditions:
\begin{itemize}
    \item \textbf{Object Counting}: For prompts requiring specific quantities (e.g., ``\textit{a photo of four handbags}'' or ``\textit{a photo of three cups}''), the baselines tend to suffer from mode collapse, either missing instances or generating redundant duplicates. In contrast, our proposed \textbf{DeMoE} accurately retains instance numerosity.
    \item \textbf{Spatial Relations}: When handling complex spatial arrangements (e.g., ``\textit{a couch below a cup}'' or ``\textit{a bird below a skateboard}''), existing methods often misplace targets or miss small conditioned entities entirely. \textbf{DeMoE} faithfully respects the conditioned spatial layouts without unnatural deformations.
    \item \textbf{Attribute and Multi-Object Binding}: In compositional scenarios containing multiple distinct entities (e.g., ``\textit{a tv and a cell phone}'' or ``\textit{a surfboard and a suitcase}''), baseline models often exhibit catastrophic attribute leakage or fail to synthesize co-occurring objects. Our approach effectively decouples multi-task representations via expert specialization, preserving sharp object boundaries and correct semantic bindings.
\end{itemize}

Overall, while single-step distillation typically degrades compositional reasoning compared to multi-step sampling, \textbf{DeMoE} not only surpasses the one-step baseline \textbf{Dense DMD2} by a substantial margin, but also consistently matches or outperforms the multi-step \textbf{Dense Teacher} in structural consistency and semantic alignment.

\subsection{Selected hard subset as a training-budget control}
\label{sec:hard-subset}

The shared MoE exceeds the original dense model by 14.58 percentage points with
paired interval $[6.25,22.92]$, and the difference against the additionally
trained dense continuation is 3.13 points with interval $[-3.13,10.42]$. Because
the subset is enriched for original-parent failures and uses a separate
evaluator port, it is a stress test rather than a population GenEval estimate.
Its role in the argument is to show that additional training alone accounts for
much of the gain on this subset, which is why the main text does not treat the
subset as a substitute for a matched full-population continuation.

\subsection{Finite route controls}
\label{sec:route-controls}

\begin{table}[H]
  \centering
  \caption{All registered route-control contrasts, in units of $10^{-3}$. Positive means
  the named control has larger recomputed pseudo-loss than the fitted route, so
  positive values favour the fitted route.}
  \label{tab:f178-contrasts}
  \footnotesize
  \begin{tabular}{lrr}
    \toprule
    Control minus fitted & Mean & Marginal 95\% interval \\
    \midrule
    Fit: natural & 0.084 & $[-0.327,\ 0.487]$ \\
    Eval: natural & 0.536 & $[0.070,\ 1.037]$ \\
    Eval: global raw & 0.957 & $[0.505,\ 1.433]$ \\
    Eval: global matched & 1.197 & $[0.733,\ 1.725]$ \\
    Eval: prompt-shuffled & 0.269 & $[-0.316,\ 0.814]$ \\
    Eval: matched-random & $-0.149$ & $[-0.703,\ 0.474]$ \\
    Eval draw 1: natural & 0.273 & $[-0.339,\ 0.799]$ \\
    Eval draw 2: natural & 0.798 & $[0.127,\ 1.594]$ \\
    \bottomrule
  \end{tabular}
\end{table}

This diagnostic freezes the persistent student, expert, tracker, and discriminator weights;
it fits prompt-specific route offsets on four draws and evaluates two
independent draws. The full control table is reproduced here because the figure
cannot be reduced to a pass/fail label: global controls are clearly worse than
the fitted route, while shuffled and matched-random controls remain unresolved.
As derived in \cref{sec:pseudoloss}, the registered scalar is a recomputed
pseudo-loss and is not upgraded to a finite KL descent statement.

\subsection{Teacher-response coordinate}
\label{sec:coordinate}

\begin{table}[H]
  \centering
  \caption{Stability of the construction-fit two-axis teacher-response
  coordinate. Each bank contains 128 prompts; the second-noise bank reuses the
  confirmation prompts.}
  \label{tab:f179-coordinate}
  \footnotesize
  \begin{tabular}{lrr}
    \toprule
    Bank & Top-two centered energy & Minimum principal cosine \\
    \midrule
    Construction & 93.124\% & 1.000000 \\
    Confirmation & 93.082\% & 0.999945 \\
    Second noise & 92.942\% & 0.999956 \\
    \bottomrule
  \end{tabular}
\end{table}

The coordinate is a stable description of four measured macro responses. It is
not evidence that two universal semantic phases exist, nor that the current
expert bank has learned them. The associated expert run contains only a
mechanics check, formal function learning and task utility remain unverified,
and the deployed method of \cref{sec:method} does not use this coordinate.

\subsection{Numerical trainability and the hard-execution switch}
\label{sec:stability}

\begin{figure}[H]
  \centering
  \includegraphics[width=0.72\linewidth]{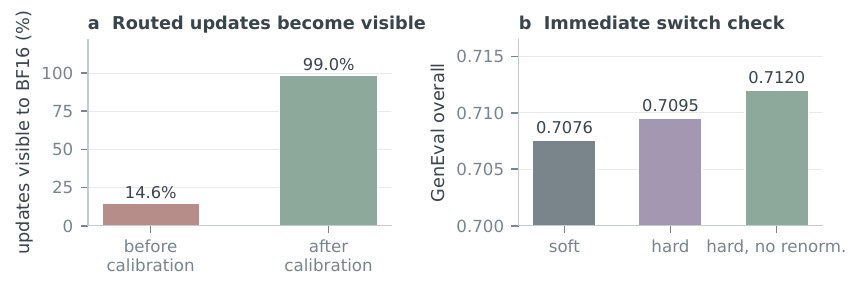}
  \caption{Numerical trainability and soft-to-hard switch checks.
  \textbf{(a)} BF16-visible update fraction and local response loss;
  \textbf{(b)} GenEval change at one earlier checkpoint. Both are
  implementation diagnostics outside the main configuration.}
  \label{fig:numerical-stability}
\end{figure}

With FP32 master updates, the fraction of updates visible to the BF16 forward
pass rises from 14.60\% to 99.02\%, and the local response loss falls from
0.24125 to 0.01860. Switching one earlier checkpoint from soft to hard
execution changes overall GenEval by less than half a point; both panels use
configurations other than the one reported in \cref{tab:main-results}. The
precision result answers whether routed experts receive usable numerical
updates under a loss that is computed by subtracting a shared output. It is not
an image-quality result and it does not explain DeMoE's main gain. The switch
check answers whether the immediate change of execution rule damaged the
checkpoint at which it was measured; it does not test whether hard execution is
better than soft execution in general. Both are reported here so that readers
can separate numerical confounds from architectural effects.

The next two tables describe the capacity and router configurations that the
main comparison is drawn from. They are design matrices: entries state what each
configuration contains, not what it achieved, and the lower block of the router
table is a prospective extension rather than a validated component.

\begin{table}[H]
\centering\small
\setlength{\tabcolsep}{4.2pt}
\renewcommand{\arraystretch}{1.10}
\caption{Expert configurations under explicit capacity control and measured 1-NFE performance.
An expert's intermediate width is $q$ times the parent FFN width, yielding total FFN ratio
$q(S+R)$ and active FFN ratio $q(S+k)$. Performance columns report text alignment (GenEval Overall),
distribution fidelity (COCO-10K FID).
Bold marks the best performance within each ablation section.}
\label{tab:expert-design}
\begin{tabular}{@{}lrrrrcc ccc@{}}
\toprule
& & & & & \multicolumn{2}{c}{FFN Width Ratio} & \multicolumn{3}{c}{1-NFE Performance} \\
\cmidrule(lr){6-7}\cmidrule(lr){8-10}
Design & Shared $S$ & Routed $R$ & Active $k$ & Width $q$
& Total & Active & GenEval $\uparrow$ & FID $\downarrow$ & Latency (s) $\downarrow$ \\
\midrule
\multicolumn{10}{@{}l}{\emph{Shared fraction: total $1\times$, active $0.5\times$ (ablation on shared capacity)}}\\
\addlinespace[2pt]
\texttt{8E4A0S}  & 0 & 8 & 4 & $1/8$  & $1$   & $1/2$ & 0.7180          & 23.4500          & 0.2185 \\
\texttt{7E3A1S}  & 1 & 7 & 3 & $1/8$  & $1$   & $1/2$ & 0.7460          & 21.1500          & 0.2142 \\
\texttt{6E2A2S}  & 2 & 6 & 2 & $1/8$  & $1$   & $1/2$ & \textbf{0.7510} & \textbf{20.8200} & 0.2124 \\
\texttt{5E1A3S}  & 3 & 5 & 1 & $1/8$  & $1$   & $1/2$ & 0.7390          & 21.8000          & \textbf{0.2096} \\
\midrule
\multicolumn{10}{@{}l}{\emph{Granularity: shared width fixed at $0.25\times$, active at $0.5\times$ (expert size vs.\ count)}}\\
\addlinespace[2pt]
\texttt{3E1A1S}  & 1 & 3 & 1 & $1/4$  & $1$   & $1/2$ & 0.7415          & 21.6266          & \textbf{0.2108} \\
\texttt{6E2A2S}  & 2 & 6 & 2 & $1/8$  & $1$   & $1/2$ & \textbf{0.7510} & \textbf{20.8200} & 0.2124 \\
\texttt{12E4A4S} & 4 & 12 & 4 & $1/16$ & $1$   & $1/2$ & 0.7485          & 21.0500          & 0.2195 \\
\midrule
\multicolumn{10}{@{}l}{\emph{Capacity scaling: widening active FFN towards full deployment}}\\
\addlinespace[2pt]
\texttt{3E1A1S}  & 1 & 3 & 1 & $1/4$  & $1$   & $0.500$ & 0.7415          & 21.6266          & \textbf{0.2108} \\
\texttt{7E3A1S}  & 1 & 7 & 3 & $1/5.4$& $1.48$& $0.740$ & 0.7812          & 16.9240          & 0.2224 \\
\texttt{7E4A1S}  & 1 & 7 & 4 & $1/6.2$& $1.29$& $0.805$ & \textbf{0.8220} & \textbf{14.8520} & 0.2256 \\
\midrule
\multicolumn{10}{@{}l}{\emph{Width compression: active expert width strictly $< 0.5\times$ (extreme efficiency)}}\\
\addlinespace[2pt]
\texttt{3E1A1S}  & 1 & 3 & 1 & $1/8$  & $1/2$ & $1/4$ & 0.7020          & 24.1500          & 0.1782 \\
\texttt{3E1A1S}  & 1 & 3 & 1 & $1/16$ & $1/4$ & $1/8$ & 0.6650          & 26.8200          & \textbf{0.1625} \\
\texttt{6E2A2S}  & 2 & 6 & 2 & $1/16$ & $1/2$ & $1/4$ & \textbf{0.7110} & \textbf{23.6400} & 0.1810 \\
\bottomrule
\end{tabular}
\end{table}

\begin{table}[H]
\centering\small
\setlength{\tabcolsep}{5pt}
\renewcommand{\arraystretch}{1.12}
\caption{Router design comparison on an identical expert bank and training
protocol. Entries record which mechanism a design contains; they are not
results. The lower block is a prospective extension and is not part of the
validated configuration.}
\label{tab:router-design}
\begin{tabular}{@{}lccccc@{}}
\toprule
Router & \shortstack{Shared\\response} & \shortstack{History\\state}
& \shortstack{Action-response\\objective} & \shortstack{Feedback\\objective}
& \shortstack{History in\\expert}\\
\midrule
Current hidden & --- & --- & --- & --- & ---\\
Current + shared & Yes & --- & --- & --- & ---\\
GRU execution observer & Yes & GRU & --- & --- & ---\\
Execution-aware TTT & Yes & Fast weights & --- & --- & ---\\
AR-CURRENT & Yes & --- & Yes & --- & ---\\
AR-HISTORY & Yes & Fast weights & Yes & --- & ---\\
\midrule
AR-CURRENT + feedback & Yes & --- & Yes & Yes & ---\\
AR-HISTORY + feedback & Yes & Fast weights & Yes & Yes & ---\\
AR-HISTORY + modulation & Yes & Fast weights & Yes & --- & Yes\\
\bottomrule
\end{tabular}
\end{table}

\section{Theory}
\label{sec:proofs}

This appendix supplies the derivations that are intentionally compressed in the
main narrative. Every statement is local to a fixed differentiable execution
branch unless stated otherwise. Crossing a hard-routing boundary or changing the
score estimator requires a separate finite experiment.

\subsection{Pathwise distribution-matching gradient}
\label{sec:pathwise-proof}

\begin{assumption}[Ideal noisy-marginal analysis]
For each condition and sampled perturbation coordinate $u$ with $\sigma_u>0$,
the student and target noisy marginals have positive differentiable densities
with sufficient integrability and boundary decay to exchange differentiation and
expectation. The target is fixed, and the score fields are exact.
\end{assumption}

Let $\bm x_0=G_\vartheta(\bm z,c)$ and
$\bm x_u=\alpha_u\bm x_0+\sigma_u\bm\epsilon$. The transported-sample variation
of $\KL(q_{\vartheta,u}\|p_{T,u})$ is the score difference $\bm s_q-\bm s_T$.
Since $\partial\bm x_u/\partial\vartheta=\alpha_uJ_\vartheta$, the chain rule
gives
\begin{equation}
  \nabla_\vartheta\KL(q_{\vartheta,u}\|p_{T,u})
  =\E\!\left[J_\vartheta^\top\alpha_u(\bm s_q-\bm s_T)\right].
\end{equation}
Multiplying by $\lambda(u)$ and averaging over $c$ and $u$ yields
\cref{eq:score-chain}. A guided neural field need not be the score of a globally
consistent density family; in that case the ideal KL interpretation is an
additional modelling assumption, while the fixed-computation-graph statements
below still apply.

\subsection{The implemented fixed-target surrogate}
\label{sec:pseudoloss}

At a reference checkpoint $\bar\vartheta$, let the implementation construct a
detached correction $\widehat{\bm\delta}^{\rm code}$ and fixed target
$\bm y=G_{\bar\vartheta}(\bm z,c)-\widehat{\bm\delta}^{\rm code}$. Then
\begin{equation}
  \ell^{\rm loc}(\vartheta;\bar\vartheta)
  =\frac{1}{2D}\norm{G_\vartheta(\bm z,c)-\sg(\bm y)}^2,
  \qquad
  \nabla_\vartheta\ell^{\rm loc}\big|_{\bar\vartheta}
  =D^{-1}J_{\bar\vartheta}^{\top}\widehat{\bm\delta}^{\rm code}.
  \label{eq:fixed-surrogate}
\end{equation}
This is exact for the frozen graph. If a candidate update recomputes both the
endpoint and its detached target, the resulting scalar is a different objective
and cannot be used as a finite-difference check of the original update. A valid
local audit freezes $\bm y$, records the actual optimiser displacement, and
evaluates endpoint and image changes separately. The route diagnostic of
\cref{sec:route-controls} uses this scalar and is therefore reported as a
pseudo-loss rather than as a certified KL quantity.

\subsection{Local interaction accounting}
\label{sec:harm-proof}

Let $a$ index a realised correction demand with nonnegative weight $w_a$, fixed
local objective $\ell_a^{\rm loc}$, and gradient $\bm g_a$. Define
$\bar{\bm g}=\sum_a w_a\bm g_a$ and $I_{ab}=\ip{\bm g_a}{\bm g_b}$.

\begin{proposition}[Signed first-order interaction]
For $\vartheta^+=\bar\vartheta-\eta\bar{\bm g}$ and twice-differentiable local
objectives,
\begin{equation}
\begin{aligned}
  \ell_a^{\rm loc}(\vartheta^+)-\ell_a^{\rm loc}(\bar\vartheta)
  =&-\eta w_a\norm{\bm g_a}^{2}
  +\eta\!\sum_{b\ne a}w_b[-I_{ab}]_+\\
  &-\eta\!\sum_{b\ne a}w_b[I_{ab}]_+ + O(\eta^2).
\end{aligned}
  \label{eq:harm-accounting}
\end{equation}
\end{proposition}

The result follows by Taylor expansion and $x=[x]_+-[-x]_+$. The three displayed
terms are self-correction, negative transfer, and positive transfer. For an
$L_a$-smooth objective the remainder is bounded by
$L_a\eta^2\norm{\bar{\bm g}}^2/2$. With AdamW, clipping, or other
preconditioning, the empirical audit must use the recorded displacement rather
than substituting a raw-gradient step.

\subsection{Exact localisation to an FFN writer}

For the second FFN projection $Y=ZW_2$, let $Q_a$ be the back-propagated
cotangent for demand $a$, including the loss reduction. Since
$\nabla_{W_2}\ell_a=Z_a^\top Q_a$,
\begin{equation}
\begin{aligned}
  I_{ab}^{W_2}
  &=\ip{Z_a^\top Q_a}{Z_b^\top Q_b}_F\\
  &=\sum_{i,j}\ip{\bm z_{ai}}{\bm z_{bj}}\ip{\bm q_{ai}}{\bm q_{bj}}.
\end{aligned}
  \label{eq:ffn-interaction}
\end{equation}
Both the activation and cotangent factors are necessary. A negative token-pair
contribution is a decomposition of a parameter-gradient interaction; it is not
by itself a semantic conflict or a causal instruction to create an expert.

\subsection{Selective separation and its cost}
\label{sec:selective-proof}

Consider a portable-gradient local model with unchanged shared gradient
$\bm s_a$, dense private gradient $\bm v_a$, and routed private gradient
$\bigoplus_e r_{ae}\bm v_a$. Let $o_{ab}=\bm r_a^\top\bm r_b$ be route overlap
and $c_{ab}=\bm v_a^\top\bm v_b$.

\begin{proposition}[When selective separation helps]
\label{prop:decouple}
The routed-minus-dense change in alignment between demand $a$ and the aggregate
update is
\begin{equation}
\begin{aligned}
\mathcal A_a^R-\mathcal A_a^D
=&\underbrace{\sum_{b\ne a}w_b(1-o_{ab})[-c_{ab}]_+}_{N_a:\ \text{negative transfer removed}}\\
&-\underbrace{\sum_{b\ne a}w_b(1-o_{ab})[c_{ab}]_+}_{P_a:\ \text{positive transfer lost}}
-\underbrace{w_a(1-o_{aa})\norm{\bm v_a}^{2}}_{D_a:\ \text{self-correction diluted}}.
\end{aligned}
  \label{eq:separation-tradeoff}
\end{equation}
\end{proposition}

The dense private interaction is $c_{ab}$ and the routed private interaction is
$o_{ab}c_{ab}$; subtracting them and separating the diagonal, positive, and
negative parts proves the result. If the compact bank changes the private
interaction by $\varepsilon_{ab}$ or alters the shared contribution by
$\varepsilon_a^S$, the positive margin must additionally exceed
$|\varepsilon_a^S|+\sum_bw_b|\varepsilon_{ab}|$. This portability error is
measured, not assumed away. Equation~\ref{eq:separation-summary} is the
main-text shorthand for \cref{eq:separation-tradeoff}.

\subsection{When can history help?}

\begin{proposition}[Conditional value of history]
\label{prop:history}
Let $A$ be a square-integrable future action-utility target, $H$ the current
deployment-visible state, and $M$ a causal summary of past executed transitions.
Under squared loss, the Bayes-risk improvement from adding history is
\begin{equation}
  \mathcal R(H)-\mathcal R(H,M)
  =\E\!\left[\operatorname{Var}\!\left(\E[A\mid H,M]\mid H\right)\right]\geq0.
  \label{eq:history-variance}
\end{equation}
It is zero exactly when $\E[A\mid H,M]=\E[A\mid H]$ almost surely.
\end{proposition}

This follows from the law of total variance. It does not guarantee that a finite
learned history model improves ranking or image quality. A valid comparison must
hold the response bank and budget fixed and include current-only, reset-history,
and donor-history controls.

\subsection{History and task-feedback extensions}
\label{sec:extensions-app}

A causal state may summarise already executed transitions,
\begin{equation}
  \bm m_{\ell+1}=U_\phi(\bm m_\ell,\bm h_\ell,K_\ell,\bm b_\ell,\bm h_{\ell+1}),
  \qquad
  \bm a_{\ell+1}=f_\psi(\bm h_{\ell+1},c,\bm m_{\ell+1}).
\end{equation}
 Its registered ranking comparison
establishes that both current and history models contain signal, but not that
history contributes beyond current hidden state.

No reviewed upstream implementation provides the exact combination of DeMoE's
executed-action record, task-grounded failure signal, and interaction-aware
redistribution. This task-feedback extension is therefore
\texttt{locally\_designed}. Its assumptions are that the failure signal is
repeatable, can influence a later legal action, leaves the bank fixed during
attribution, and transfers to image quality. Candidate alternatives include a
current-only residual controller, contextual-bandit credit assignment, dense
reranking, and transport-constrained allocation. The extension is falsified if
it cannot beat a matched current-only control, if reset or donor history
performs equally well, or if proxy improvements do not change the generated-image
metric.

\begin{figure}[p]
  \centering
  \includegraphics[page=18, width=\linewidth, height=0.42\textheight, keepaspectratio]{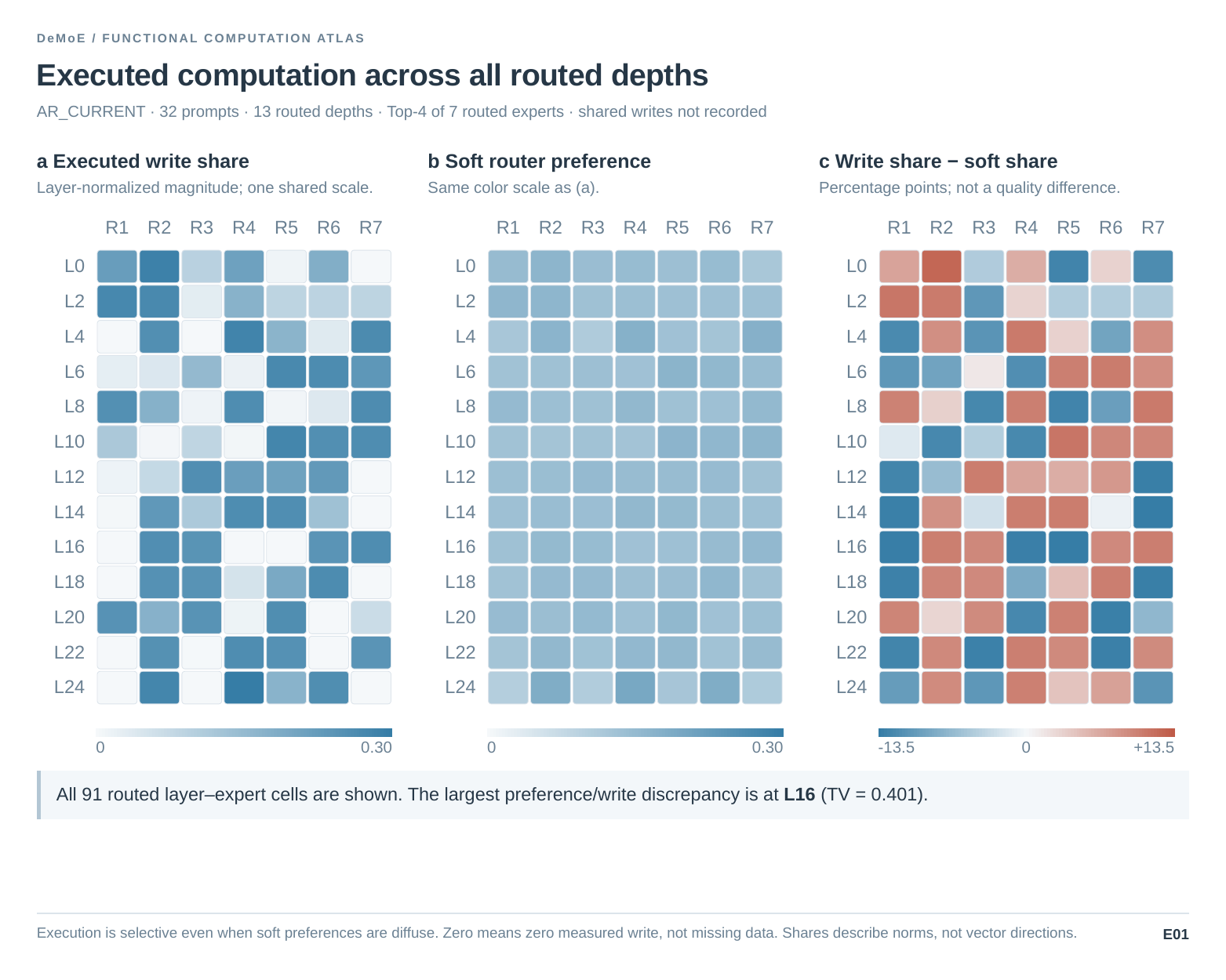}
  \par\vspace{1ex}
  \includegraphics[page=19, width=\linewidth, height=0.42\textheight, keepaspectratio]{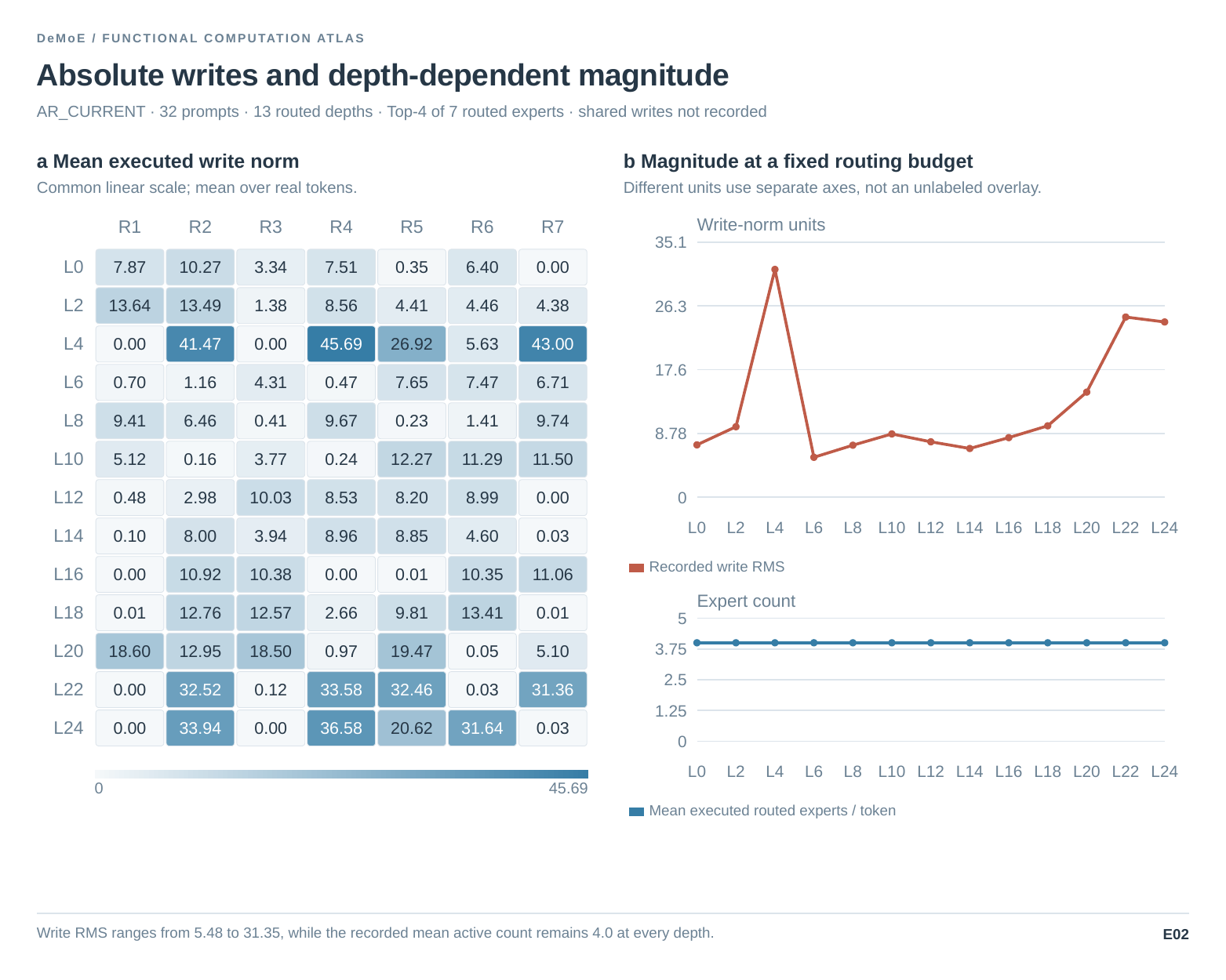}
  \caption{Executed write shares and soft preferences at all 13 routed depths.}
  \label{fig:atlas-executed-writes}
\end{figure}

\begin{figure}[p]
  
  \centering
  \includegraphics[page=20, width=\linewidth, height=0.42\textheight, keepaspectratio]{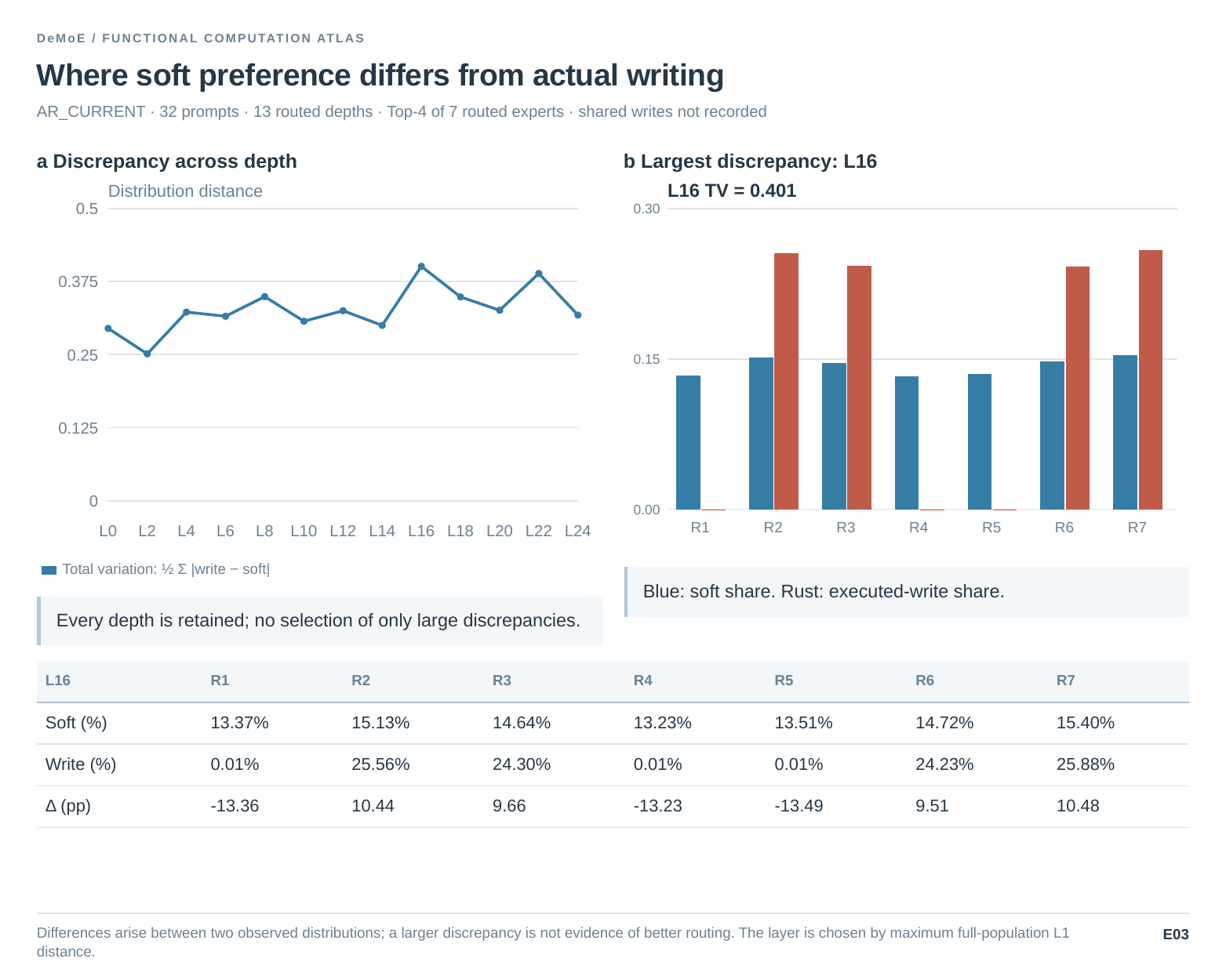}
  \par\vspace{1ex}
  \includegraphics[page=21, width=\linewidth, height=0.42\textheight, keepaspectratio]{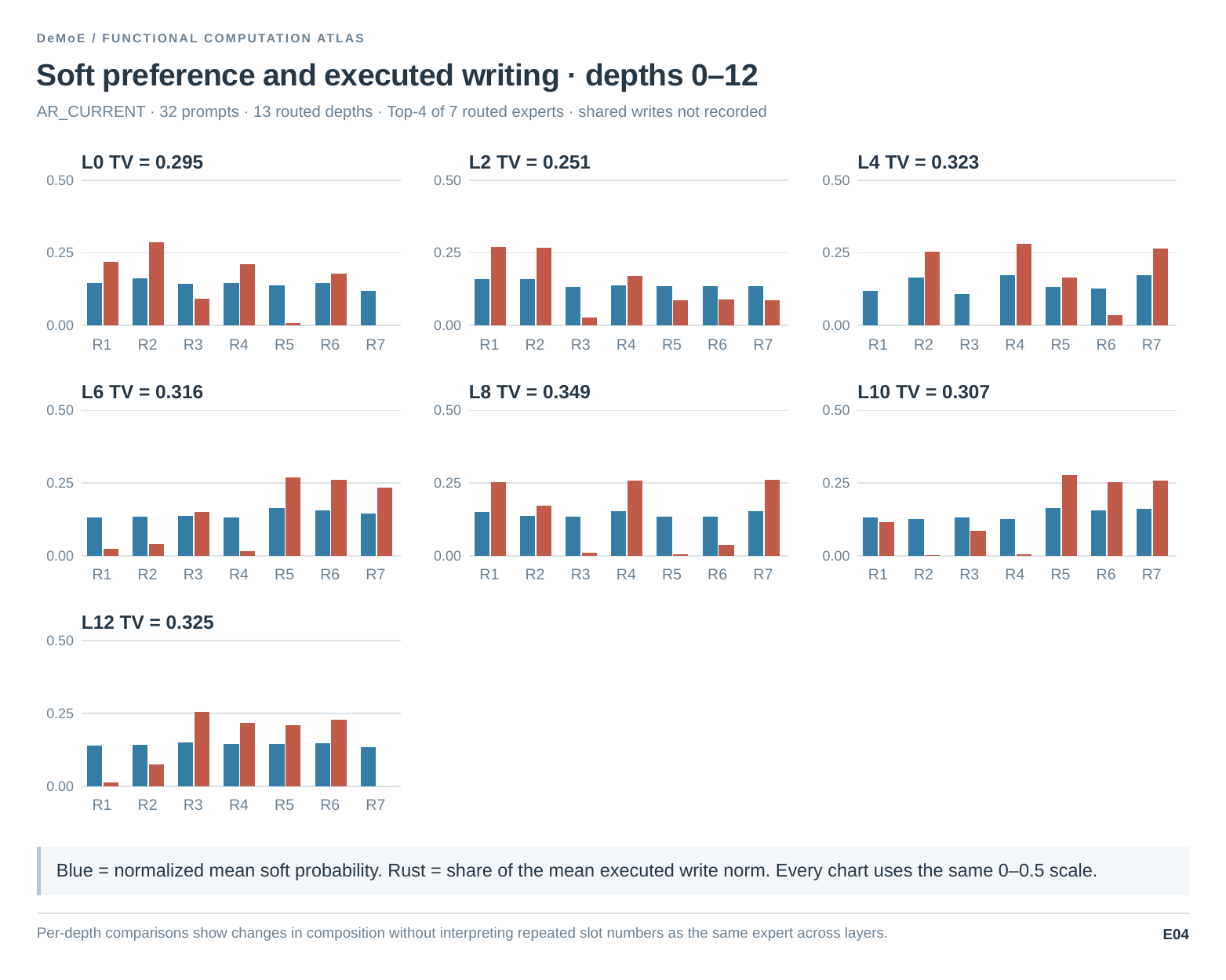}
  \caption[]{Executed write shares and soft preferences (continued).}
\end{figure}

\begin{figure}[p]
  
  \centering
  \includegraphics[page=22, width=\linewidth, height=0.42\textheight, keepaspectratio]{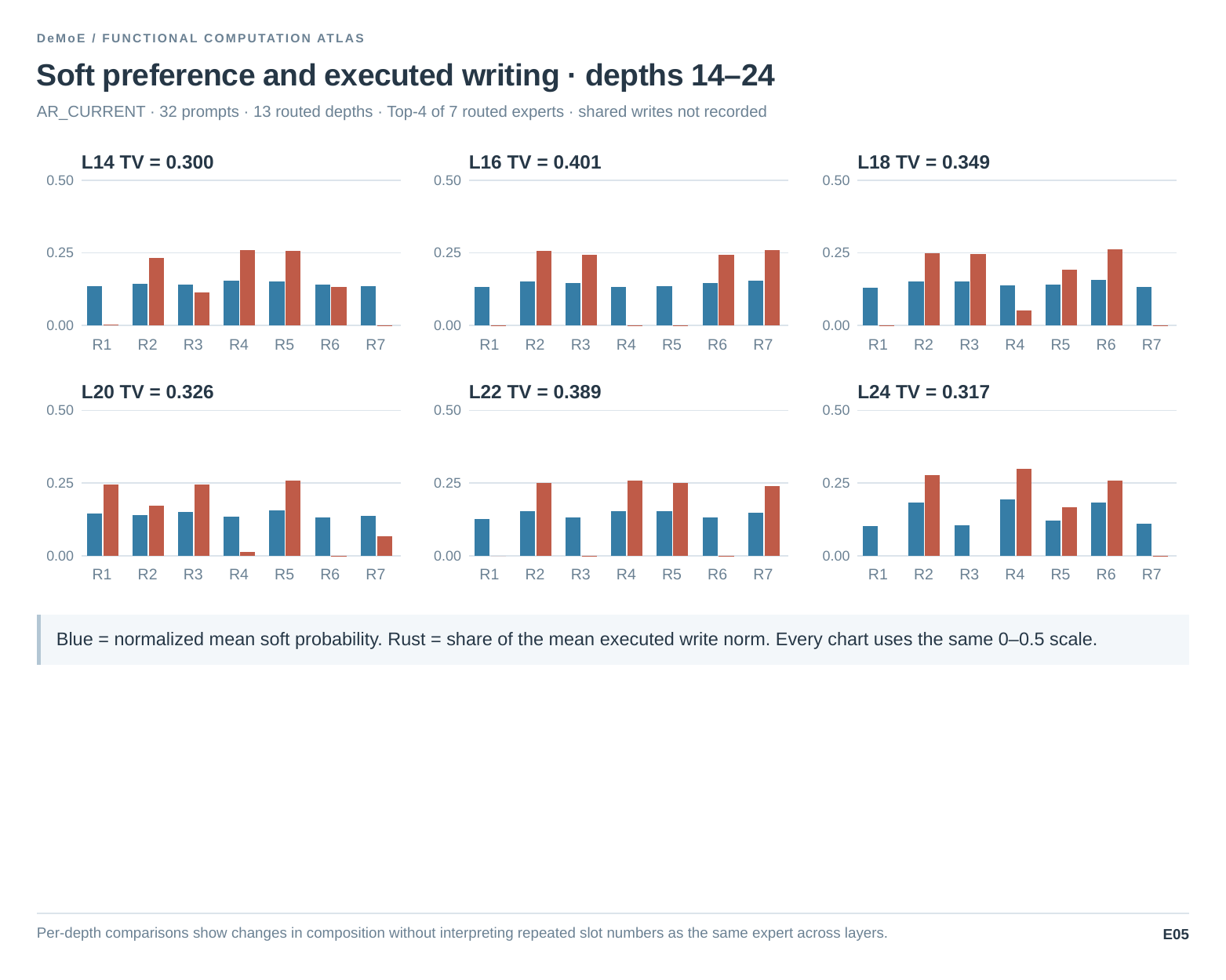}
  \par\vspace{1ex}
  \includegraphics[page=23, width=\linewidth, height=0.42\textheight, keepaspectratio]{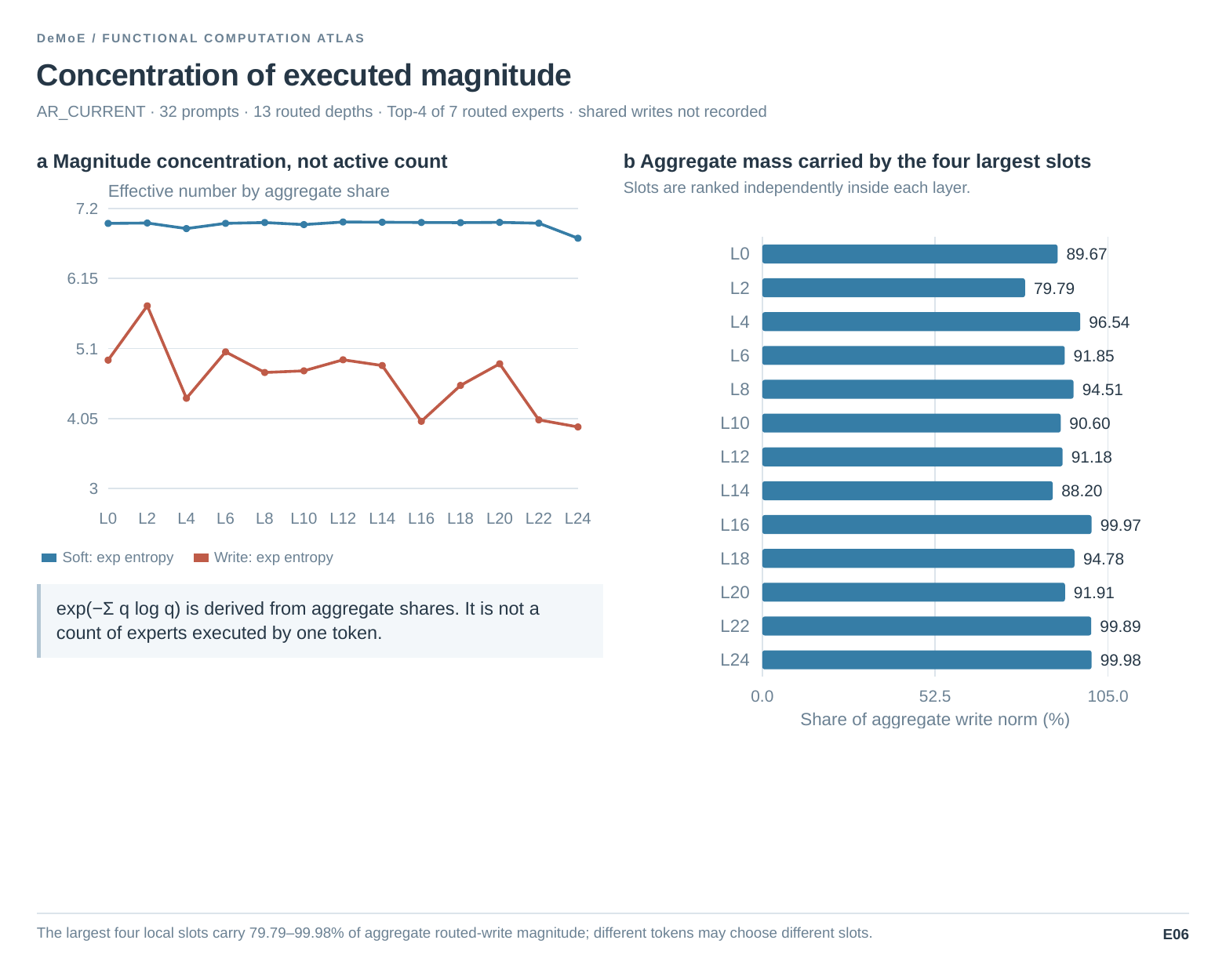}
  \caption[]{Executed write shares and soft preferences (continued).}
\end{figure}

\begin{figure}[p]
  
  \centering
  \includegraphics[page=24, width=\linewidth, height=0.42\textheight, keepaspectratio]{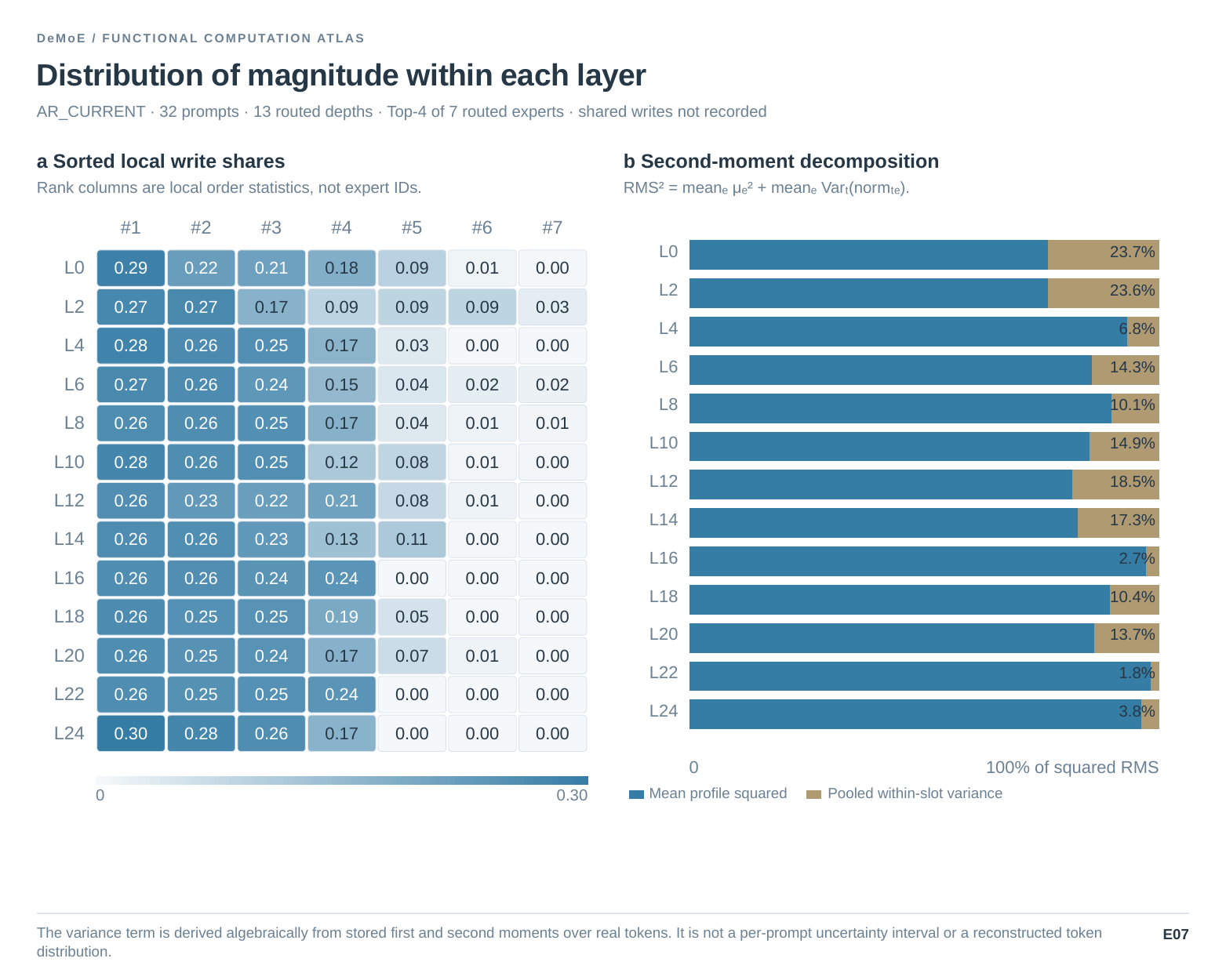}
  \par\vspace{1ex}
  \includegraphics[page=25, width=\linewidth, height=0.42\textheight, keepaspectratio]{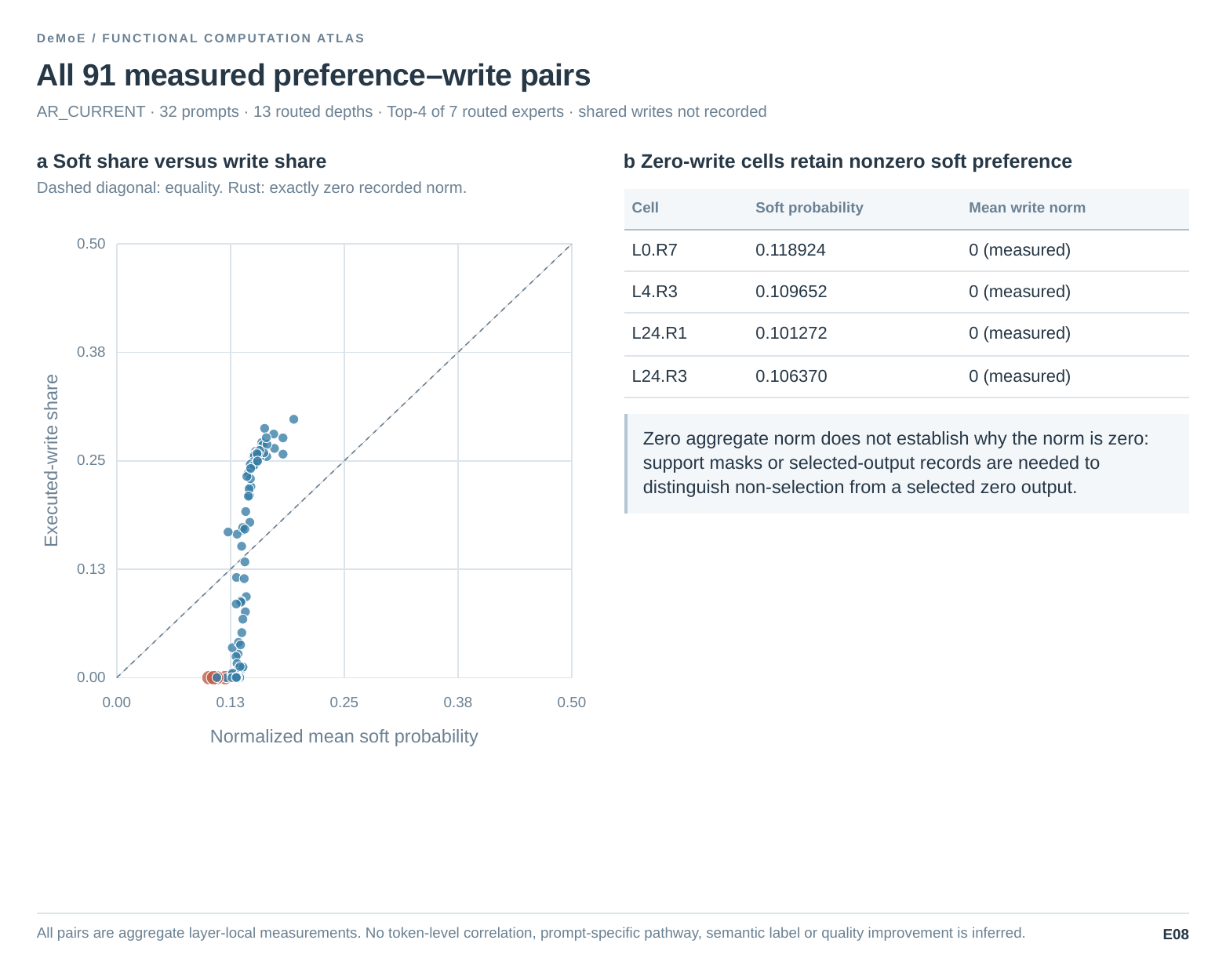}
  \caption[]{Executed write shares and soft preferences (continued).}
\end{figure}

\begin{figure}[p]
  
  \centering
  \includegraphics[page=26, width=\linewidth, height=0.42\textheight, keepaspectratio]{figures/DeMoE_Atlas_All_Figures.pdf}
  \par\vspace{1ex}
  \includegraphics[page=1, width=\linewidth, height=0.42\textheight, keepaspectratio]{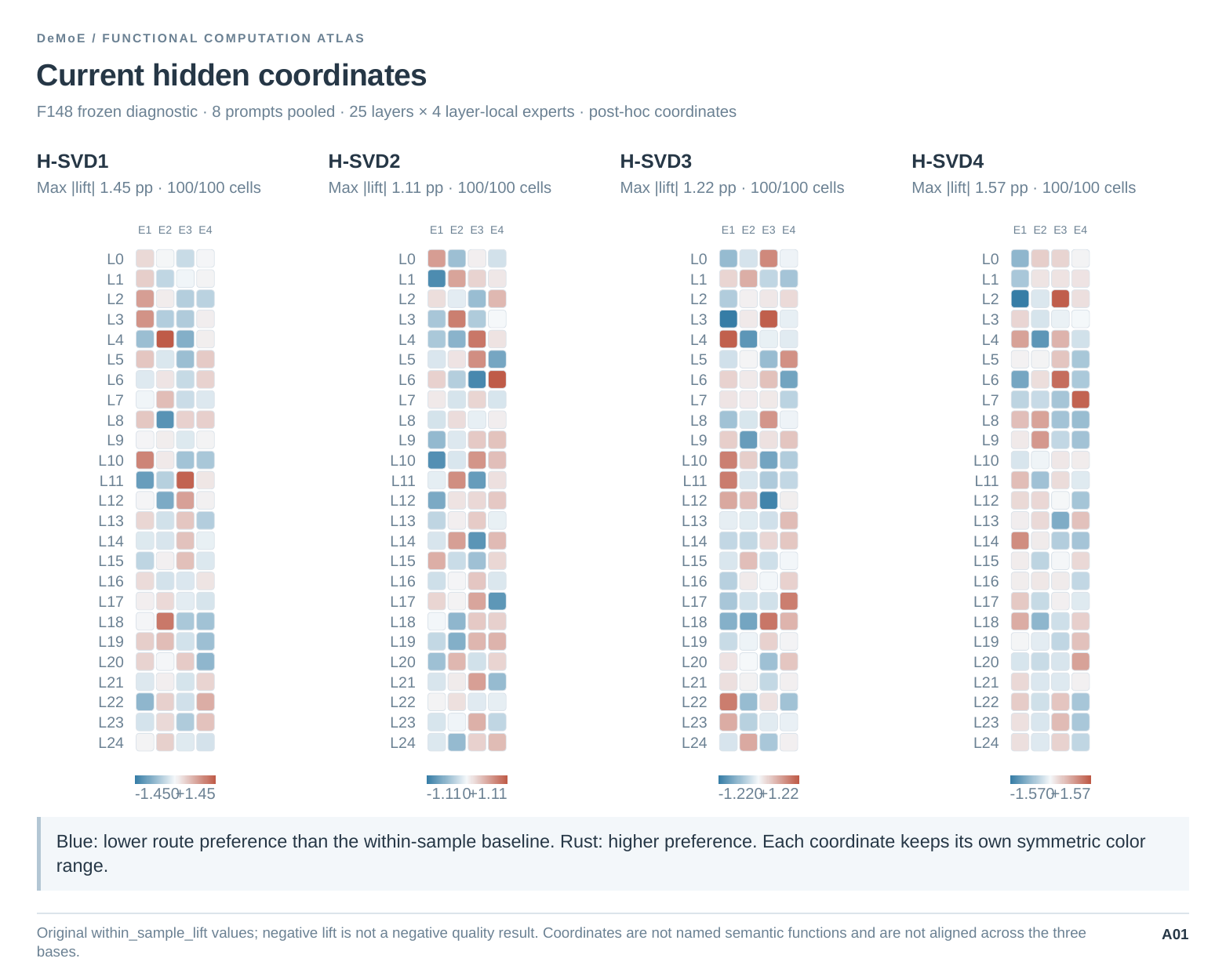}
  \caption[]{Executed write shares and soft preferences (continued).}
\end{figure}

\begin{figure}[p]
  
  \centering
  \includegraphics[page=11, width=\linewidth, height=0.42\textheight, keepaspectratio]{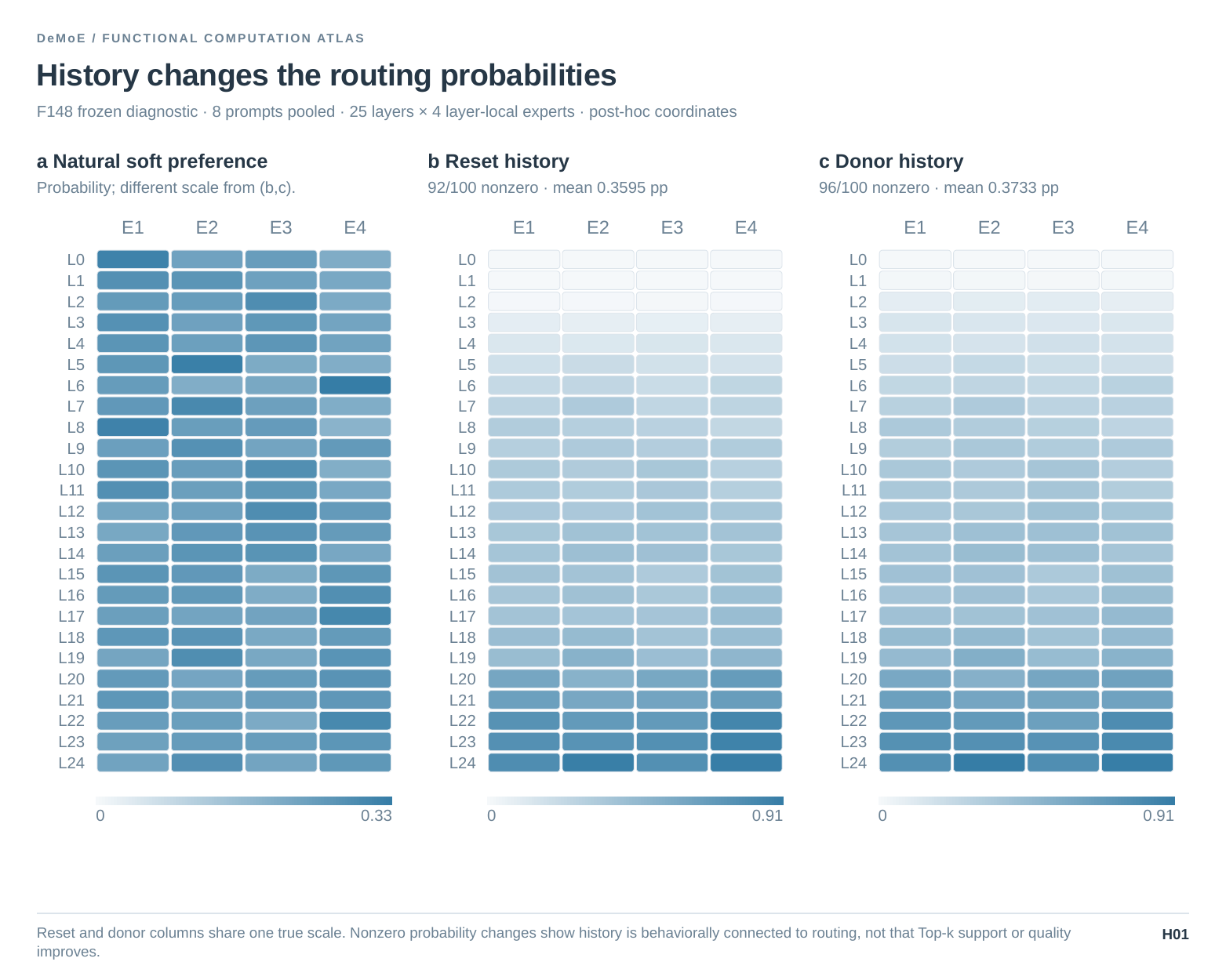}
  \par\vspace{1ex}
  \includegraphics[page=12, width=\linewidth, height=0.42\textheight, keepaspectratio]{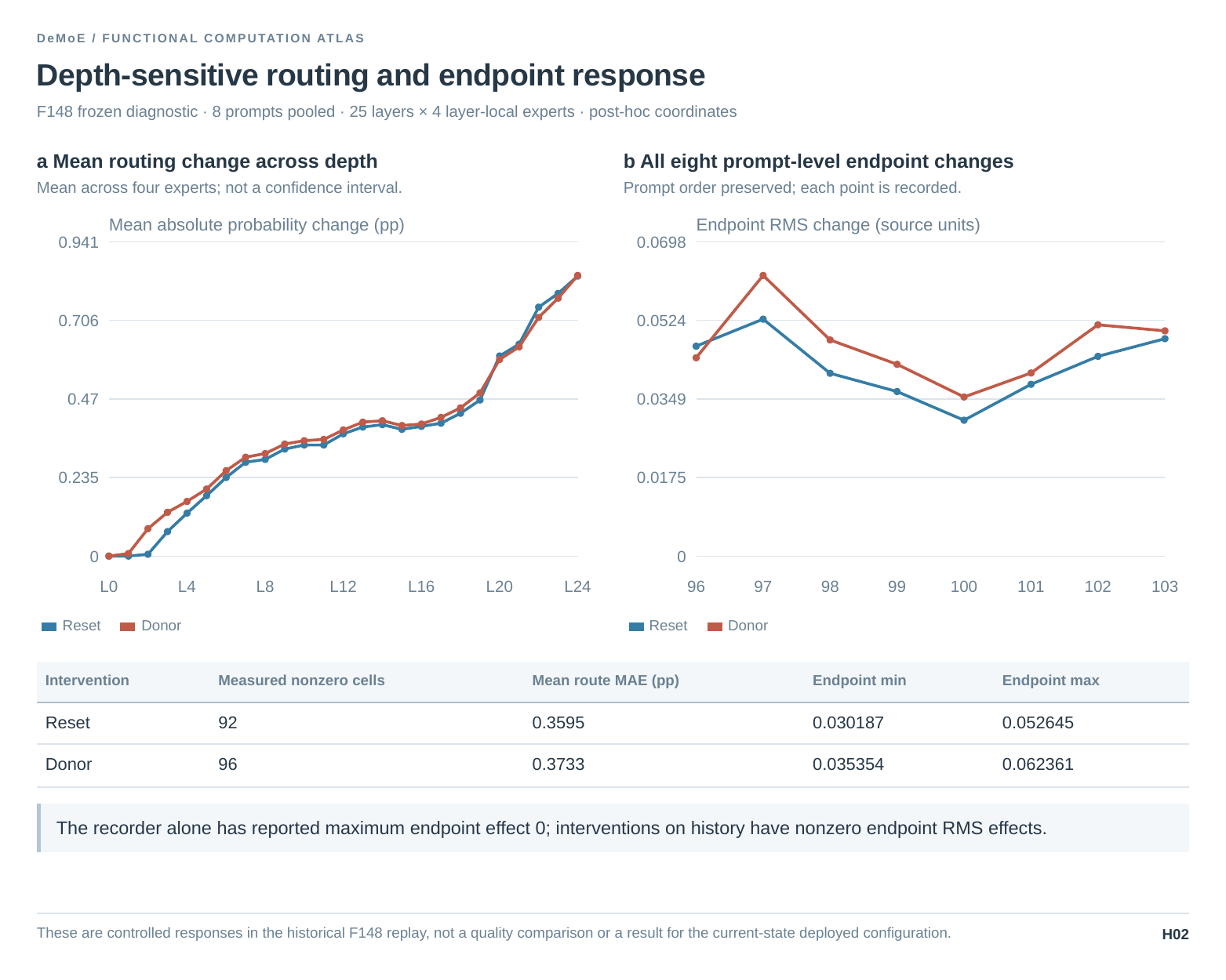}
  \caption[]{Executed write shares and soft preferences (continued).}
\end{figure}

\begin{figure}[p]
  
  \centering
  \includegraphics[page=16, width=\linewidth, height=0.42\textheight, keepaspectratio]{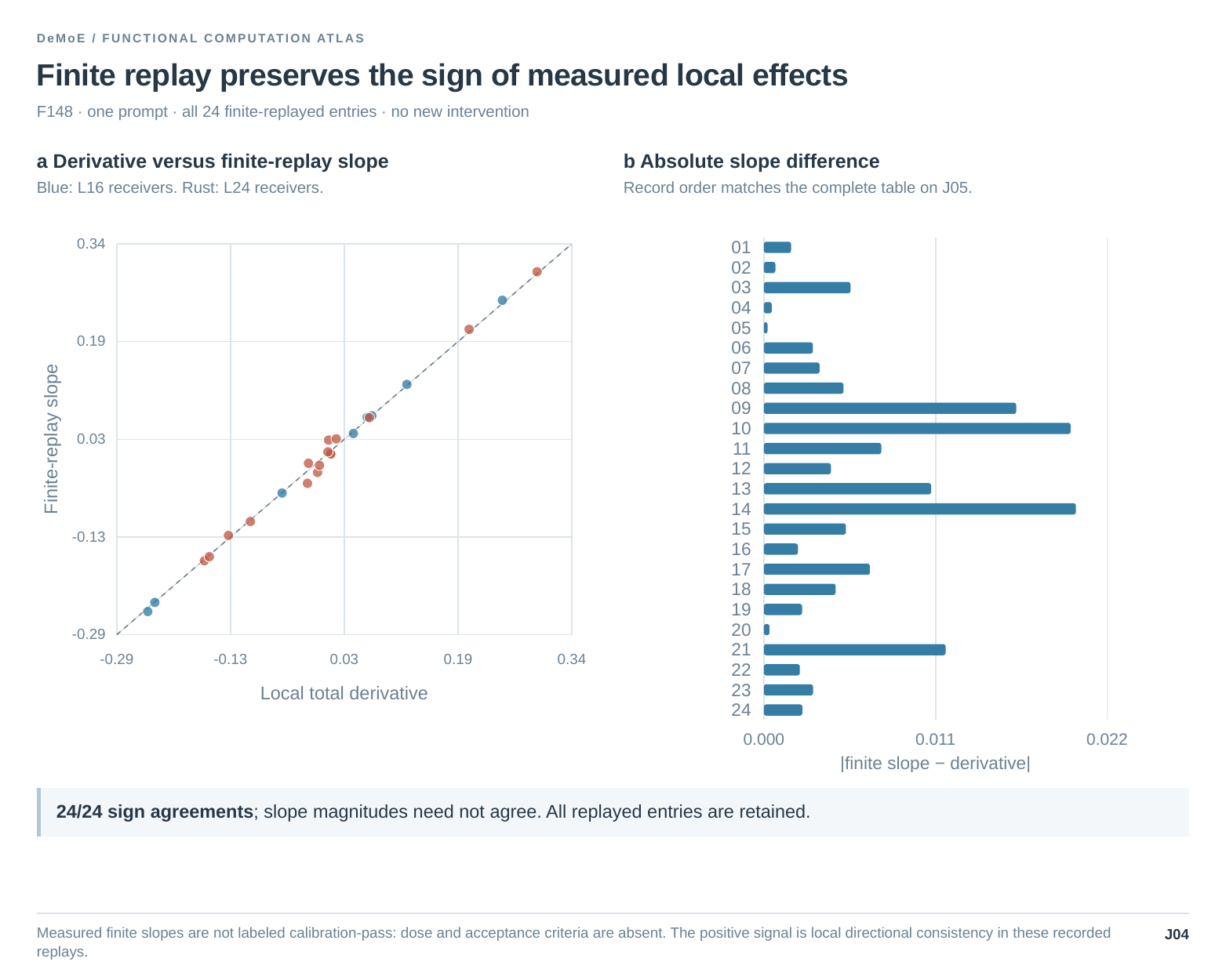}
  \par\vspace{1ex}
  \includegraphics[page=17, width=\linewidth, height=0.42\textheight, keepaspectratio]{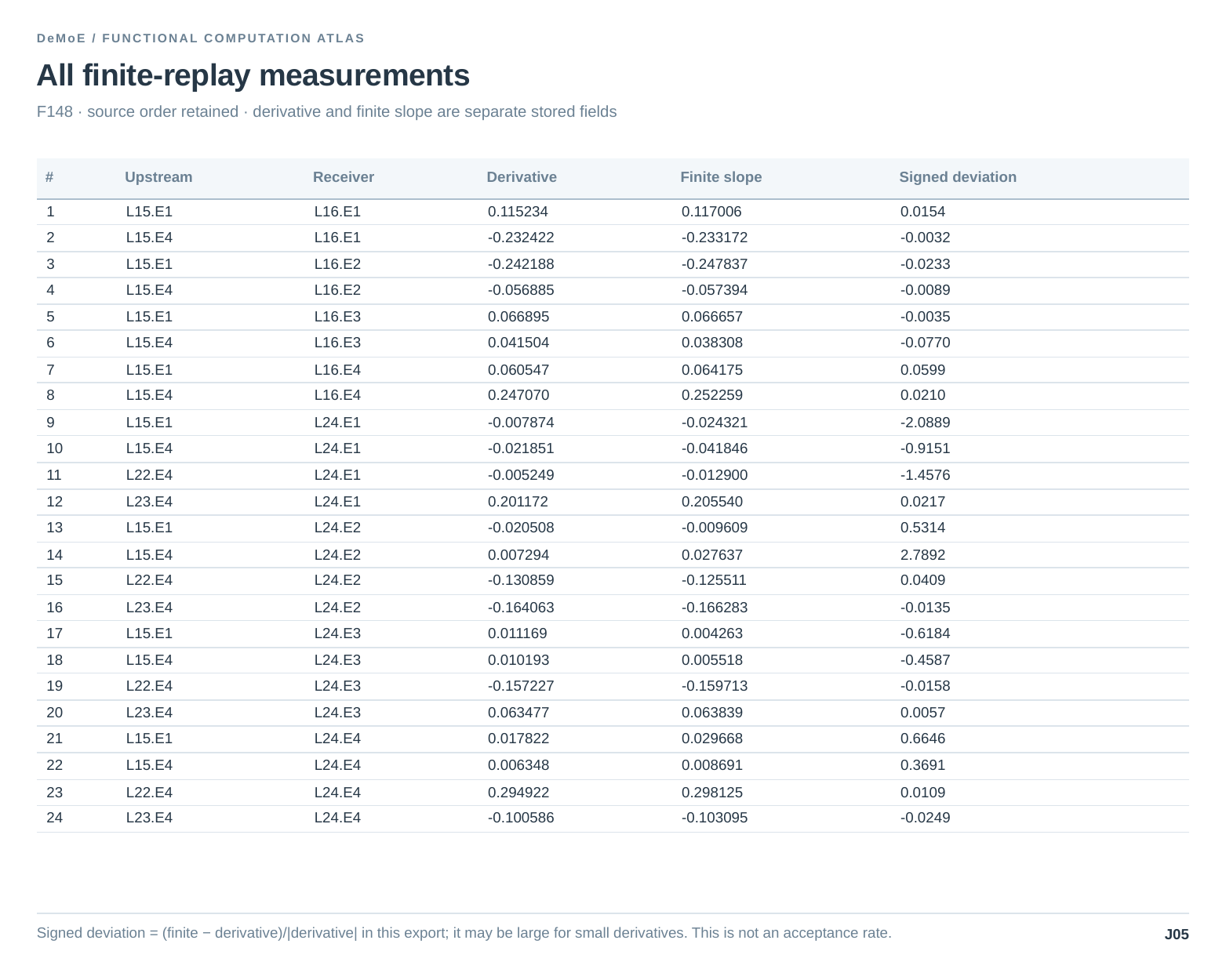}
  \caption[]{Executed write shares and soft preferences (continued).}
\end{figure}

\section{Extended Related Work}
\label{sec:related}

\paragraph{Sparse experts and diffusion routing.}
Sparse mixture-of-experts (MoE) architectures dynamically route tokens to decouple parameter capacity from inference compute~\citep{shazeer2017outrageously,fedus2022switch}. Shared-expert paradigms mitigate routing instability by maintaining universal functional paths~\citep{dai2024deepseekmoe}. In diffusion models, MoE has been adapted to partition timesteps, tokens, or noise regimes across denoiser sub-networks~\citep{fei2024scaling,sun2024ec,park2023dtr,park2024switchdit,fang2024remixdit,shi2025diffmoe,Cheng2025DiffMoE,zheng2025dense2moe,wei2025routing,yuan2025expert}. Beyond vision synthesis, diffusion MoE architectures serve as expressive multimodal action policies in robotics and reinforcement learning~\citep{zhou2024vdd,reuss2025mode,hao2026smp}. Unlike multi-step routing along external diffusion timesteps~\citep{Fan2025PhasedDMD,Wang2025DDT}, DeMoE governs computation purely across internal layer depth within a single forward pass.

\paragraph{Sparse experts and shared computation.}
Sparse mixture-of-experts layers make parameter paths conditionally
executable~\citep{shazeer2017outrageously,fedus2022switch}, and the motivation
there is capacity and conditional computation per token. Shared-expert designs
deliberately retain common computation while routed experts
specialise~\citep{dai2024deepseekmoe}. Diffusion-transformer work has routed
tokens~\citep{fei2024scaling,sun2024ec,park2024switchdit,fang2024remixdit},
denoising tasks, and training phases across experts. DeMoE reuses the structural
idea of splitting a feed-forward width into an always-active part and a
selectable part. It differs in the responsibility it attaches to the routing
decision inside a one-step student: the executed support is recorded and the
response that actually ran becomes the supervised target of the auxiliary
prediction loss.

\paragraph{Sparse experts and diffusion routing.}
Sparse mixture-of-experts (MoE) architectures dynamically route tokens to decouple parameter capacity from inference compute~\citep{shazeer2017outrageously,fedus2022switch}. Shared-expert paradigms mitigate routing instability by maintaining universal functional paths~\citep{dai2024deepseekmoe}. In diffusion models, MoE has been adapted to partition timesteps, tokens, or noise regimes across denoiser sub-networks~\citep{fei2024scaling,sun2024ec,park2023dtr,park2024switchdit,fang2024remixdit,shi2025diffmoe,Cheng2025DiffMoE,zheng2025dense2moe,wei2025routing,yuan2025expert}. Beyond vision synthesis, diffusion MoE architectures serve as expressive multimodal action policies in robotics and reinforcement learning~\citep{zhou2024vdd,reuss2025mode,hao2026smp}. Unlike multi-step routing along external diffusion timesteps~\citep{Fan2025PhasedDMD,Wang2025DDT}, DeMoE governs computation purely across internal layer depth within a single forward pass.

\paragraph{Gradient interaction and negative transfer.}
Multi-task methods diagnose or modify conflicting parameter
gradients~\citep{yu2020pcgrad}, and diffusion models can exhibit negative
transfer across denoising tasks~\citep{go2023negative}. Our local analysis uses
the same mathematical object, a signed inner product between parameter
gradients, but it does not infer a universal causal chain from reverse-KL
training to conflict and then to mode loss. The selective-separation
proposition of \cref{sec:selective-proof} explicitly retains positive transfer,
self-correction, and a parameter-geometry error term, and the empirical support
for the design is the controlled-divergence and allocation evidence of
\cref{sec:allocation}.

\paragraph{Cost accounting for conditional computation.}
Conditional execution reduces the parameters touched by a given token, but the
realised benefit depends on kernel shapes, dispatch overhead, and memory
traffic, which is why sparse layers are a capacity mechanism before they are an
efficiency guarantee~\citep{shazeer2017outrageously,fedus2022switch}. DeMoE
follows the same discipline: the active-width ratio in
\cref{tab:main-results} is a structural description of the configuration, and we
therefore make no latency, memory, energy, or total-FLOP claim, because no
matched measurement exists.

\paragraph{Mode collapse, diversity preservation, and particle guidance.}
A persistent vulnerability in score distillation and one-step compression is mode collapse and diversity loss~\citep{wang2024esd,cai2026mmm,sargent2025flowmo}. Remedies include incorporating explicit diversity-preserving regularizers into distillation objectives~\citep{wu2026diversitypreserved}. Complementary sampling-stage approaches incorporate inter-particle repulsion and particle guidance to preserve generative variance~\citep{corso2024particle,vinograd2026eddy,sadat2024cads,zilberstein2025repulsive,blessing2025gmp}, alongside dual-expert and swift distillation designs~\citep{li2026duet,nguyen2024swiftbrush}.
\end{document}